\documentclass[11pt]{article}

\usepackage[preprint]{acl}

\usepackage{times}
\usepackage{latexsym}

\usepackage[T1]{fontenc}

\usepackage[utf8]{inputenc}

\usepackage{microtype}

\usepackage{inconsolata}

\usepackage{graphicx}
\usepackage[most]{tcolorbox}
\usepackage{enumitem}
\usepackage{booktabs}
\usepackage{tabularx}
\usepackage{multirow}
\usepackage{multicol}
\usepackage{ragged2e}
\usepackage[table]{xcolor}
\usepackage{pifont}
\setlist[itemize]{noitemsep,nolistsep}
\usepackage{subcaption}

\usepackage{CJKutf8}
\usepackage[utf8]{inputenc}

\DeclareUnicodeCharacter{0259}{\rotatebox[origin=c]{180}{e}} 

\usepackage[english]{babel}
\usepackage[colorinlistoftodos]{todonotes} 

\newcolumntype{L}{>{\raggedleft\arraybackslash}X} 
\newcolumntype{Y}{>{\raggedright\arraybackslash}X} 
\newcolumntype{F}{p{4em}}

\newcommand{\borderlines}{\textsc{BorderLines}}

\newcommand{\csmetric}{\texttt{CS}}
\newcommand{\kbcs}{\texttt{KB\_CS}}
\newcommand{\cstcs}{\texttt{Cst\_CS}}
\newcommand{\deltacs}{\texttt{$\Delta$\_CS}}
\newcommand{\concs}{\texttt{Con\_CS}}
\newcommand{\noncs}{\texttt{Non\_CS}}

\newcommand{\noir}{\texttt{no\_ir}}
\newcommand{\qlang}{\texttt{qlang}}

\newcommand{\rellangs}{\texttt{rel\_langs}}

\definecolor{carmine}{HTML}{960018}
\definecolor{royalblue}{HTML}{0038A8}

\newcommand{\smip}{\textcolor{royalblue}{IP}}
\newcommand{\smis}{\textcolor{royalblue}{IS}}
\newcommand{\smca}{\textcolor{carmine}{Ca}}
\newcommand{\smsc}{\textcolor{carmine}{SC}}

\newcommand{\smindep}{\textcolor{royalblue}{Ind}}
\newcommand{\smstate}{\textcolor{carmine}{Cnt}}

\newcommand{\dataset}{\textsc{\textcolor{OliveGreen}{WorldStateMedia}}}
\newcommand{\dataseten}{\dataset\textcolor{OliveGreen}{\textsubscript{en}}}
\newcommand{\datasetmulti}{\dataset\textcolor{OliveGreen}{\textsubscript{multi}}}

\title{The World as Viewed by 136 Countries: An Open, Curated, Annotated, and Multilingual Dataset of 118 Million State Media Articles}

\author{
Bryan Li$^{1}$\thanks{Contact: \texttt{bryanli.ca@gmail.com}.
Work done while at the University of Pennsylvania.
$^\dagger$Equal second contribution},
Samar Haider$^{3\dagger}$,
Grace Kim$^{2\dagger}$,
Fiona Luo$^4$,
Jinqi Luo$^2$, \\
\textbf{Yuan Yuan$^5$, Duncan Watts$^2$,
Ren\'e Vidal$^2$, Chris Callison-Burch$^2$}
\\
$^{1}$Google \enspace
$^{2}$University of Pennsylvania \enspace
$^{3}$Northwestern University \\
$^{4}$University of California, Berkeley \enspace
$^{5}$University of California, Los Angeles
}

\begin{document}
\maketitle
\begin{abstract}
State media outlets often project their governments' narratives to target both domestic and foreign audiences. Because this content is plentiful and free to access, it appears widely in the web corpora that large language models (LLMs) are trained on and that retrieval systems draw from. However, the prevalence of state media in these corpora and its potential influence on LLM outputs remain largely unexplored. In this paper, we introduce \dataset, an open dataset of 118 million state media articles from 136 countries in 114 languages. We extract these articles from existing pretraining corpora and annotate each outlet with level of state control from State Media Monitor. Using \dataset\ to audit these corpora, we find that, in English web crawls, the share of state media pages from editorially controlled (vs. independent) outlets rose from 39\% in 2013 to 93\% in 2025. In several low-resource languages, state media make up a substantial share of all web data, such as 17.8\% for Uyghur and 7.8\% for Belarusian. To characterize the differing narratives of news outlets, we compare Russian and Ukrainian state media in English and in their native languages, revealing asymmetric topical focus and opposing framings of contested issues such as Crimea. Furthermore, in a retrieval-augmented generation (RAG) experiment on territorial disputes, retrieving a country's state media shifts model answers toward that country's claims, most strongly for English-language (foreign-targeted) media. We release \dataset\ to support multilingual, cross-national insights on state media.\footnote{Dataset, code, and artifacts at: \url{https://huggingface.co/collections/manestay/worldstatemedia}}

\end{abstract}

\section{Introduction}
\label{sec:intro}

Large language models (LLMs) learn from, and now also retrieve from, the open web. The contents of web corpora shape models' understanding about the world, and a growing body of work audits these corpora for their sources, quality, and consent~\citep{dodge-etal-2021-documenting,kreutzer-etal-2022-quality,elazar2024wimbd,longpre2024consent}. One such source is \textit{state media}: news outlets owned, operated, or heavily influenced by a national government, thus projecting those governments' authority and geopolitical stances~\citep{neveu2004government}. 

State media outlets span a spectrum of control from editorially independent public broadcasters such as the BBC, which aim to provide a shared factual reference point~\citep{cushion2017democratic}, to state-controlled outlets in authoritarian countries such as RT, which serve to monopolize information flow~\citep{walker2014breaking}. Governments use state media to promote strategic narratives among two primary audiences, domestic citizens and the international community ~\citep{miskimmon2013strategic}. In the Russia--Ukraine war, for example, Russian media emphasizes de facto territorial control, whereas Ukrainian media emphasizes internationally recognized de jure control. Because state media is typically free to access, published at high volume, and available in many languages, its narratives are well positioned to enter the corpora that LLMs are trained on and retrieve from.

Prior work on political bias in LLMs has largely focused on partisan leaning within domestic U.S. or European political contexts, e.g., how left- or right-leaning training data shifts model behavior~\citep{feng-etal-2023-pretraining,chen-etal-2024-susceptible,bang-etal-2024-measuring}. State media instead raise a different question of how the narratives of multiple governments---in many languages and aimed at different audiences---are represented in LLM training corpora data and influence their responses~\citep{waight2026state}. However, there currently exists no open, web-scale study that has systematically linked Internet crawl data to outlet-level state control to examine the prevalence of state media in LLM training corpora, their distribution across countries and languages, their degree of editorial independence, and their intended domestic or international audiences.

To answer these questions, we introduce \dataset, a large-scale open dataset of 118 million state media news articles from 136 countries in 114 languages. It is assembled by processing the Fineweb web crawl corpora~\citep{penedo2024fineweb1,penedo2024fineweb2} and extracting all news articles tracked in the State Media Monitor database~\citep{dragomir2021state}.
To demonstrate the utility of \dataset\ in helping the research community better understand the influence of state media on the behavior of AI systems in contexts from training to retrieval, we address the following research questions:

\begin{itemize}
    \item \textbf{How much state media is in LLM web data?} (\S\ref{sec:analysis}) Within English crawls, the share of state media pages from editorially controlled outlets rose from 39\% in 2013 to 93\% in 2025. In several low-resource languages, state media makes up a substantial share of all web data, up to 17.8\% for Uyghur.
    \item \textbf{What do these narratives say?} (\S\ref{sec:case-study}) A case study of Russian and Ukrainian state media, using decontextualized fact extraction and information pursuit, shows that Russia's English-language state media covers international affairs broadly while Ukraine's focuses on Ukrainian affairs, and that the two express opposing stances on Crimea.
    \item \textbf{Does retrieval over state media influence LLM responses?} (\S\ref{sec:rag}) We study a retrieval-augmented generation (RAG) scenario over different countries' state media on the \borderlines\ dataset of territorial disputes~\cite{li2024land}. We highlight several vulnerabilities, including how retrieval over foreign-targeted state media especially amplifies divergent cross-lingual responses by RAG systems.
\end{itemize}
\section{Related Work}
\label{sec:related}

\paragraph{State Media and Strategic Political Narratives}
State media has been studied as a means for governments to shape domestic and international political narratives~\citep{miskimmon2013strategic,walker2014breaking,chubb2023authoritarian}. Domestically, state-run outlets manage the visibility of information~\citep{wedeen2015ambiguities,roberts2018censored}, and exposure to state media has been associated with reduced protest and collective action~\citep{carter2021propaganda}. Governments also combine political messages with entertainment and other non-political content to reach broader audiences~\citep{mattingly2022soft,baggott2024broadcasting,zhu2024propaganda,han2025propaganda}. Internationally, state media presents governments' perspectives on geopolitical events~\citep{cook2022beijing}, and experimental evidence shows that exposure can influence foreign audiences' perceptions of a government's legitimacy and competence~\citep{mattingly2025chinese}. State media narratives can also spread through social media, including across national borders~\citep{hanley2023happenstance,hanley2025across}. These findings raise questions about how state media narratives may also influence AI systems trained on or retrieving from news content.

\paragraph{Political Bias in LLMs}
Prior work has shown that political biases in training data can influence language model behavior. \citet{feng-etal-2023-pretraining} find that pretrained language models reflect political biases in their training corpora, which propagate to downstream tasks such as hate speech and misinformation detection. \citet{chen-etal-2024-susceptible} show that even small amounts of ideologically biased instruction-tuning data can shift model positions, with effects generalizing across topics. More directly, \citet{waight2026state} identify Chinese state-coordinated media in multilingual training corpora and show that continued pretraining on this content makes models more favorable toward the Chinese government. Other studies find that LLM responses to territorial disputes vary across languages~\citep{li2024land}, including when using multilingual retrieval~\citep{li-etal-2025-multilingual}. Multilingual retrieval systems are also shown to favor evidence in the query language or high-resource languages~\citep{wu2024not,park2025investigating,sharma-etal-2025-faux}, while news sources in different languages can present different perspectives on the same events~\citep{akyurek-etal-2020-multi,piskorski-etal-2023-multilingual}. Our work extends these studies by identifying state media in LLM pretraining corpora across 136 countries and 114 languages, analyzing differences in their narratives, and examining how retrieving from these sources affects LLM responses.

\begin{table*}[t]
\centering
\scriptsize
\setlength{\tabcolsep}{5pt}

\begin{tabularx}{\textwidth}{
>{\raggedright\arraybackslash}X
c
c
c
c
c
c
p{1.2cm}
c
}
\toprule
\textbf{Dataset} & \textbf{Languages} & \textbf{Size} & \textbf{Countries} & \textbf{Years} & \textbf{Full Text} & \textbf{Deduped} & \textbf{Influence Focused} & \textbf{License} \\
\midrule

\rowcolor{gray!12}
\dataset & \textbf{114} & \textbf{118M art.} & \textbf{136} & \textbf{2013--25} & \textbf{Yes} & \textbf{Yes} & \textbf{Yes} & \textbf{ODC-By} \\

\midrule
\multicolumn{9}{c}{\textit{Large Multilingual News Corpora}} \\
\midrule

GDELT \textsuperscript{\cite{Leetaru13gdelt:global}}
& 100+ & $\sim$2.5TB/yr & 230+ & 1979-- & Yes & No & No & Unrestricted \\

CC-News\textsuperscript{\cite{commoncrawl}}
& 100+ & $\sim$1.35B art. & 220+ & 2016-- & Yes & No & No & CC ToU \\

Media Cloud\textsuperscript{\cite{roberts2021media}}
& 20+ & 1.7B+ art. & Global & 2010-- & Mixed & Yes & No & API \\

Event Registry\textsuperscript{\cite{leban2014event}}
& 60+ & 150k src. & Global & 2014-- & Yes & Yes & No & Commercial \\

\midrule
\multicolumn{9}{c}{\textit{Propaganda / Influence Datasets}} \\
\midrule

PTC\textsuperscript{\cite{da2019fine}}
& 1 & 536 art. & US/West. & 2019--20 & Yes & Yes & Yes & CC BY 4.0 \\

RRN/WoF\textsuperscript{\cite{heppell-etal-2023-analysing}}
& 7 & 14k art. & Global & 2022--23 & Yes & Yes & Yes & Open \\

NELA-GT-2020\textsuperscript{\cite{gruppi2021nelagt2020largemultilabellednews}}
& 1 & 1.78M art. & Global (EN) & 2020 & Yes & Yes & No & Harvard Dataverse \\

\bottomrule
\end{tabularx}


\caption{
Comparison of multilingual news and propaganda datasets with \dataset. Existing multilingual news corpora provide broad global coverage but lack focus on influence, and have more restrictive licenses. Existing propaganda datasets are more limited in size and language coverage.
}
\label{tab:dataset_comparison}
\end{table*}
\paragraph{Comparison with Related Datasets} 
The extensive news and propaganda literature includes many resources. In Table~\ref{tab:dataset_comparison}, we compare our dataset to prior ones, showing that \dataset\ provides broad global coverage, influence focus, and a permissive license.

\section{Dataset Collection}
\label{sec:collection}
We build \dataset\ by first identifying state media websites and their level of state control (\S\ref{sec:collection_smm}), then extracting their pages from LLM pretraining corpora (\S\ref{sec:collection_articles}).

\subsection{State Media Outlets and Websites}
\label{sec:collection_smm}

\paragraph{State Media Monitor}
We draw state media outlets and their state control labels from State Media Monitor (SMM), an actively updated database of state and public media worldwide~\citep{dragomir2021state}. SMM assesses each media company's funding, ownership and governance, and editorial control, and assigns it one of seven levels of control, from most to least controlled:\footnote{\url{https://statemediamonitor.com/typology}} \textcolor{carmine}{1) State Controlled, 2) Captured Public, 3) Captured Private}, \textcolor{royalblue}{4) Independent State Funded and Managed, 5) Independent State Funded, 6) Independent State Managed, 7) Independent Public}.
In this work, we condense the first three labels into one State Controlled label (\smstate), and the latter four levels into one Independent label (\smindep). We do this for ease of analysis of broader trends for independent vs. controlled state media; the released \dataset\ dataset contains the granular seven labels. As SMM is updated annually, we apply the 2025 typology to all outlets and pages.\footnote{We acknowledge that levels of state control change over time for specific outlets, and that back-projection of 2025 levels has some limitations. This is discussed further in \S~\ref{sec:analysis_time}.}

\begin{table*}[t!]
    \tiny
    \centering

    \begin{tabularx}{\textwidth}{F Y Y F F Y}
    \toprule
    Country & Media company & Main assets & Typology & Region & URL\_dict \\
    \midrule
    Argentina & Radio y Televisión Argentina (RTA) & \textit{Television}: Televisión Pública (Canal 7), Canal 12; \textit{Radio}: Radio Nacional & \smsc & South America & \{Televisión Pública (Canal 7) : tvpublica.com.ar , Radio Nacional : radionacional.com.ar\} \\
    China & Xinhua News Agency & \textit{News agency}: Xinhua; \textit{Publishing}: Reference News (Cānkǎo Xiāoxī); \textit{Television}: CNC World & \smsc & East Asia & \{Xinhua : xinhuanet.com/english , Reference News (Cānkǎo Xiāoxī) : cankaoxiaoxi.com\} \\
    South \newline Korea & Yonhap News Agency & \textit{News agency}: Yonhap; \textit{Television}: Yonhap News TV & \smip & East Asia & \{Yonhap : en.yna.co.kr Yonhap News TV : www.yonhapnewstv.co.kr\} \\
    France & France Médias Monde & \textit{Radio}: Radio France Internationale (RFI), Monte Carlo Doualiya (MCD) \textit{Television}: France24 & \smis & Western Europe & \{Radio France Internationale (RFI) : rfi.fr/fr , Monte Carlo Doualiya (MCD) : mc-doualiya.com , France24 : france24.com/fr\} \\
    Kenya & Kenya News Agency (KNA) & \textit{News agency}: KNA & \smsc & Eastern Africa & \{KNA : kenyanews.go.ke\} \\
    Qatar & Al Jazeera Media Network (AJMN) & \textit{Television}: Al Jazeera Arabic, Al Jazeera English & \smsc & Middle East & \{Al Jazeera Arabic : aljazeera.net , Al Jazeera English : aljazeera.com\} \\
    Serbia & Ta\v{c}no & \textit{News agency}: Tanjug & \smca & Central Europe & \{Tanjug :  http://www.tanjug.rs\}\\
    Russia & Public Television of Russia (OTR) & \textit{Television}: OTR & \smsc & Russia & \{OTR : otr-online.ru\} \\
    USA & U.S. Agency for Global Media (USAGM) & \textit{Radio}: Voice of America, Radio Free Europe/Radio Liberty, Radio Free Asia, Office for Cuba Broadcasting & \smsc & North America & \{Voice of America : voanews.com , Radio Free Europe/Radio Liberty : rferl.org , Radio Free Asia : rfa.org/english , Office for Cuba  Broadcasting : martinoticias.com\} \\
    \bottomrule
\end{tabularx}

    \caption{Excerpted entries from the state media websites table, along with annotations. Although our analyses use binary \smindep/\smstate\ labels, here we show the full 7 label typology from SMM.}
    \label{tab:smm}
\end{table*}

\paragraph{Mapping outlets to web domains}
SMM lists each state outlet's media assets, e.g., television, radio, or newspapers. We map companies to their web domains in two passes. First, we prompt an LLM to fill in URLs for each asset in context (prompts in Appendix~\ref{sec:prompt_urls}). Second, we manually verify whether each URL is correct and valid; if not, we attempt to search for the correct one. The resulting table has 640 state media outlets, 1,910 assets, and 1,183 unique URLs from 173 countries. Example entries are shown in Table~\ref{tab:smm}.

\subsection{State Media Articles}
\label{sec:collection_articles}
We extract articles from existing LLM pretraining corpora instead of crawling outlets directly, so that \dataset\ reflects the state media content that models are trained on. We use the FineWeb corpora. English FineWeb (v1.4.0) covers Common Crawl dumps from mid 2013 to mid 2025, with over 15T tokens of English text~\citep{penedo2024fineweb1}. FineWeb-2 (v2.1.1) covers mid 2013 to mid 2024, with over 3T words in 1,868 language-script pairs~\citep{penedo2024fineweb2}. Both corpora apply high quality and fully reproducible filtering and deduplication pipelines.

The two corpora are organized into subsets differently: by crawl date for English FineWeb, and by language for FineWeb-2. Either way, we extract from every subset the pages where the domains match any of the state media URLs. To additionally ensure good coverage and size in our final dataset, we keep only countries with over 10k pages, and languages with over 100k pages from any source. This results in 136 countries and 114 languages (\S\ref{sec:analysis}). Appendix Table~\ref{tab:sample_articles} provides samples of articles along with their metadata.

\begin{table*}[t!]
    \scriptsize
    \centering
    \setlength{\tabcolsep}{2.11pt}
    \begin{tabularx}{\textwidth}{X r L c c r L c c | X r L c c r L c c}
\toprule
 & \multicolumn{4}{c}{English} & \multicolumn{4}{c|}{Multilingual} & & \multicolumn{4}{c}{English} & \multicolumn{4}{c}{Multilingual} \\
\cmidrule(lr){2-5} \cmidrule(lr){6-9} \cmidrule(lr){11-14} \cmidrule(lr){15-18}
Country & Rank & Pages & \smindep & \smstate & Rank & Pages & \smindep & \smstate & Country & Rank & Pages & \smindep & \smstate & Rank & Pages & \smindep & \smstate \\
\midrule
USA & 1 & 15.7m & 41.7 & 58.3 & 36 & 208k & 0.4 & 99.6 & France & 14 & 1.36m & 100.0 & 0.0 & 5 & 2.08m & 100.0 & 0.0 \\
Russia & 2 & 8.26m & 0.0 & 100.0 & 1 & 6.22m & 0.0 & 100.0 & Egypt & 15 & 1.09m & 0.0 & 100.0 & 30 & 296k & 0.0 & 100.0 \\
Australia & 3 & 6.67m & 100.0 & 0.0 & 72 & 35k & 100.0 & 0.0 & Cambodia & 16 & 1.09m & 0.0 & 9.6 & 52 & 82k & 0.0 & 100.0 \\
UK & 4 & 6.35m & 100.0 & 0.0 & 53 & 81.3k & 100.0 & 0.0 & Germany & 17 & 1.02m & 100.0 & 0.0 & 3 & 2.49m & 100.0 & 0.0 \\
Canada & 5 & 4.98m & 100.0 & 0.0 & 26 & 363k & 100.0 & 0.0 & Switzerland & 23 & 466k & 100.0 & 0.0 & 11 & 900k & 100.0 & 0.0 \\
Singapore & 6 & 4.61m & 0.0 & 100.0 & 25 & 386k & 0.0 & 100.0 & Hungary & 25 & 293k & 0.0 & 100.0 & 7 & 1.53m & 0.0 & 100.0 \\
S. Africa & 7 & 3.75m & 0.0 & 100.0 & 85 & 21.3k & 0.0 & 100.0 & Spain & 40 & 122k & 14.5 & 85.5 & 9 & 1.18m & 17.6 & 75.4 \\
Qatar & 8 & 3.58m & 0.0 & 100.0 & 14 & 803k & 0.0 & 100.0 & Poland & 43 & 109k & 78.1 & 0.0 & 4 & 2.34m & 95.0 & 0.0 \\
Ireland & 9 & 2.65m & 100.0 & 0.0 & 104 & 9,817 & 100.0 & 0.0 & Belarus & 52 & 76.6k & 0.0 & 100.0 & 12 & 852k & 0.0 & 100.0 \\
China & 10 & 2.35m & 0.0 & 100.0 & 20 & 419k & 0.0 & 100.0 & Serbia & 53 & 74.7k & 0.0 & 100.0 & 10 & 991k & 0.0 & 100.0 \\
Turkey & 11 & 2.3m & 0.0 & 100.0 & 2 & 3.42m & 0.0 & 93.5 & Finland & 57 & 62.3k & 100.0 & 0.0 & 16 & 522k & 100.0 & 0.0 \\
UAE & 12 & 2.02m & 0.0 & 97.9 & 13 & 833k & 5.7 & 62.3 & Sweden & 60 & 51.8k & 100.0 & 0.0 & 6 & 1.65m & 100.0 & 0.0 \\
Vietnam & 13 & 1.68m & 0.0 & 100.0 & 8 & 1.22m & 0.0 & 100.0 & Norway & 109 & 2,091 & 100.0 & 0.0 & 15 & 692k & 100.0 & 0.0 \\
\bottomrule
\end{tabularx}
    \caption{Dataset statistics for the 26 countries that rank in the top 16 of either the English or multilingual split of \dataset\ (136 countries total). For each country and split, we show the ordinal rank, the number of pages, and the percentage of pages from either independent (\smindep) or controlled (\smstate) state media.}
    \label{tab:fw_stats_top16}
\end{table*}

\section{Dataset Analysis}
\label{sec:analysis}

Across the English and multilingual splits, \dataset\ contains 118m articles from 136 countries in 114 languages.\footnote{Fineweb is deduplicated within each dump, rather than across the corpora, and \dataset\ inherits this. If we were to deduplicate at the corpus-level, then there are 58m articles; however, because \dataset\ is used as a lens into LLM training corpora, 118m articles is accurate. More discussion in Appendix~\ref{sec:app_dedup}.} We analyze both splits by time, language, and relative country rankings, focusing on the top 16 countries in Table~\ref{tab:fw_stats_top16}. This serves as an audit of the state media content present in LLM training corpora. Statistics over all 136 countries are given in Appendix Table~\ref{tab:fw_stats}.

\subsection{English split}
Referring to Table~\ref{tab:fw_stats_top16}, the USA has the most pages (15.7m); 58.3\% of which come from \smstate\ outlets, and 41.7\% from \smindep.\footnote{This is notably due to a re-classification of the outlet U.S. Agency for Global Media (USAGM) from \smindep\ to \smstate\ in 2025. 2024 and earlier would have been 100\% \smindep.} Russia is second, with 8.26m pages all from \smstate\ outlets.  Besides Anglophone countries, the top 16 includes China, Vietnam, Cambodia, Qatar, and the UAE. Their English-language state media not only reaches international audiences, but also increasingly serves to influence AI systems in both training and retrieval corpora.

\begin{figure}[t!]
    \centering
    \includegraphics[width=.9\linewidth]{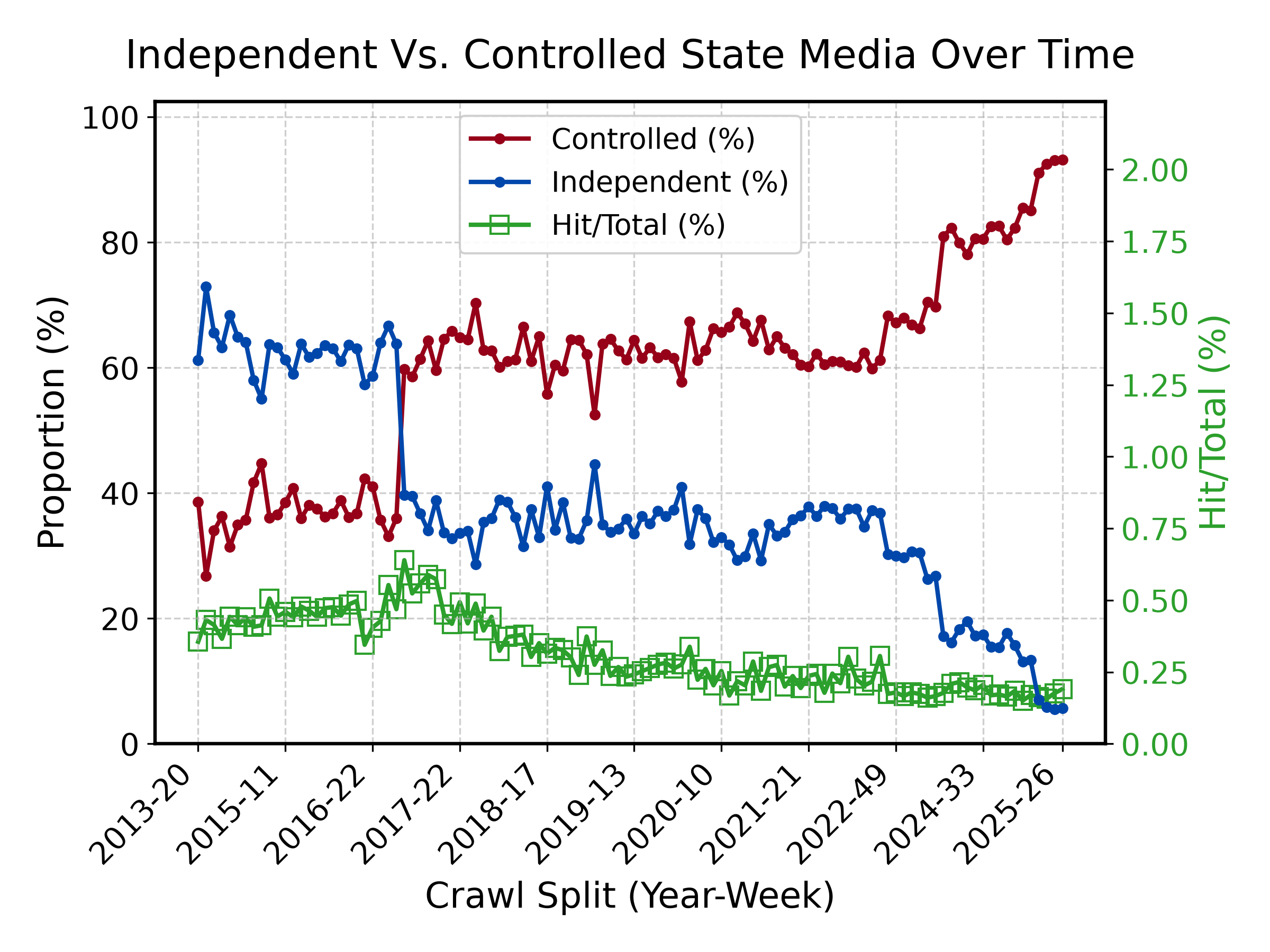}
    \caption{For \dataseten, we graph three proportions over time: within state media pages, from controlled (\smstate) vs. independent (\smindep) sources; and state media pages over all pages.}
    \label{fig:fw_en_date_pages}
\end{figure}

\paragraph{Webpages over time}
\label{sec:analysis_time}
Referring to Figure~\ref{fig:fw_en_date_pages}, we see a demonstrable shift from independent to controlled state media. We see three periods, interrupted by two abrupt shifts. In the first period, mid 2013 to mid 2016, most state media pages are \smindep\ at \textcolor{royalblue}{60\%}, and \smstate\ pages \textcolor{carmine}{40\%}. In the second period, late 2016 to mid 2024, it is inverted to \smindep\ at \textcolor{royalblue}{40\%}, and \smstate\ pages \textcolor{carmine}{60\%}. In the third period, late 2024 to mid 2025, there is a sharp climb of \smstate\ to \textcolor{carmine}{93\%} by the last crawl date.\footnote{While we apply the 2025 SMM typology to all years, we acknowledge state control can change over time, as discussed in Limitations. However, the 2013--2025 range when applying 2022 SMM is 33\% to 82\% -- in line with 39\% to 93\% above.}

By considering the exact page counts of \smstate\ and \smindep\ for each crawl in \dataseten, we identify likely causes for each shift. The 2016 shift is caused by more \smstate\ content, as \smstate\ pages more than doubling, while \smindep\ pages hold steady. The 2024 shift, meanwhile, is caused by publishers opting out of web crawlers, as \smindep\ pages fall several-fold while \smstate\ pages change far less. 

In the same figure, the green line shows the prevalence of state media pages as a proportion of all pages. This is low and stays low: from 0.36\% in 2013 to 0.19\% in 2025. There is a notable rise to 0.6\% in early 2017, before a gradual drop-off. The decrease over time is likely due to two factors: a sheer increase in the overall web content, and news publishers opting out, which we have found above to be more the case for \smindep\ ones.

\subsection{Multilingual split} 
Referring to Table~\ref{tab:fw_stats_top16}, the top countries fall into two groups: largely \smstate\ with Russia, Turkey, Hungary, Vietnam, Serbia, and Belarus, and largely \smindep\ with Germany, Poland, France, Sweden, Switzerland, Norway, and Finland. Notable population outliers from the \smstate\ group include Serbia (rank 10) with 991k pages for 6.6m people, and Belarus (rank 12) with 852k pages for 9m people.

\subsection{By language}
\label{sec:analysis_lang}
For each language, we compute the share of all FineWeb-2 pages that come from state media. Table~\ref{tab:fwmulti_lang_top} shows the 14 highest.

Several low-resource languages have high shares, mostly from \smstate\ outlets: Uyghur (17.8\%), Turkmen (8.3\%), Belarusian (7.8\%), Serbian in Latin script (7.0\%), and Khmer (6.8\%). Others have similar shares from \smindep\ public broadcasters, such as Norwegian Nynorsk (7.7\%), Hausa (7.0\%), M\=aori (6.8\%), and Scottish Gaelic (6.4\%).
Among higher-resource languages, Arabic (5.3\%) and Turkish (3.4\%) have notable shares, mostly \smstate. Russian and Chinese have many state media pages, but other web content dilutes their shares to 1.0\% and 0.2\%.
Web data in low-resource languages is known to be scarce and often low in quality~\citep{kreutzer-etal-2022-quality}. We add that in several such languages, the outlets of one or a few states supply 5--18\% of it. For these languages, state media is a prominent source in both training data and retrieval pools.

\paragraph{English vs. multilingual split rankings}
Comparing a country's rank in the two splits (English/multilingual) suggests different orientations. Russia (\#2/\#1) and China (\#10/\#20) rank high in both, consistent with state media that addresses both domestic and international audiences. So do Qatar (\#8/\#14) and the UAE (\#12/\#13), given their small populations. Turkey (\#11/\#2), France (\#14/\#5), Germany (\#17/\#3), and Hungary (\#25/\#7) rank much higher in the multilingual split, consistent with a domestic focus. However, these observations on differing focuses of state media come with the caveat that they may be partially attributable to differences in crawl coverage and filtering for the source datasets FineWeb and FineWeb-2.

\begin{table}[t]
    \footnotesize
    \centering
    \setlength{\tabcolsep}{3pt}
\begin{tabular}{l r r | l r r}
\toprule
Lang. & Hit \% & \smstate & Language & Hit & \smstate \\
\midrule
Uyghur & 17.8 & 58.5 & M\=aori & 6.8 & 0.0 \\
Turkmen & 8.3 & 62.8 & Amharic & 6.5 & 43.7 \\
Belar. & 7.8 & 59.9 & Sc.\ Gaelic & 6.4 & 0.0 \\
Norweg. & 7.7 & 0.0 & Pashto & 5.9 & 31.2 \\
Serbian (Lat.) & 7.0 & 95.5 & Arabic & 5.3 & 83.3 \\
Hausa & 7.0 & 2.7 & Kazakh & 5.3 & 68.0 \\
Khmer & 6.8 & 85.1 & Serbian (Cyr.) & 5.1 & 99.9 \\
\bottomrule
\end{tabular}

    \caption{The 14 languages in \datasetmulti\ with the highest proportion of pages coming from state media outlets (Hit \%), and of those the proportion of those from \smstate\ outlets.}
    \label{tab:fwmulti_lang_top}
\end{table}

\begin{figure}[t]
    \centering
    \begin{subfigure}[t]{0.49\textwidth}
        \centering
        \includegraphics[width=\linewidth]{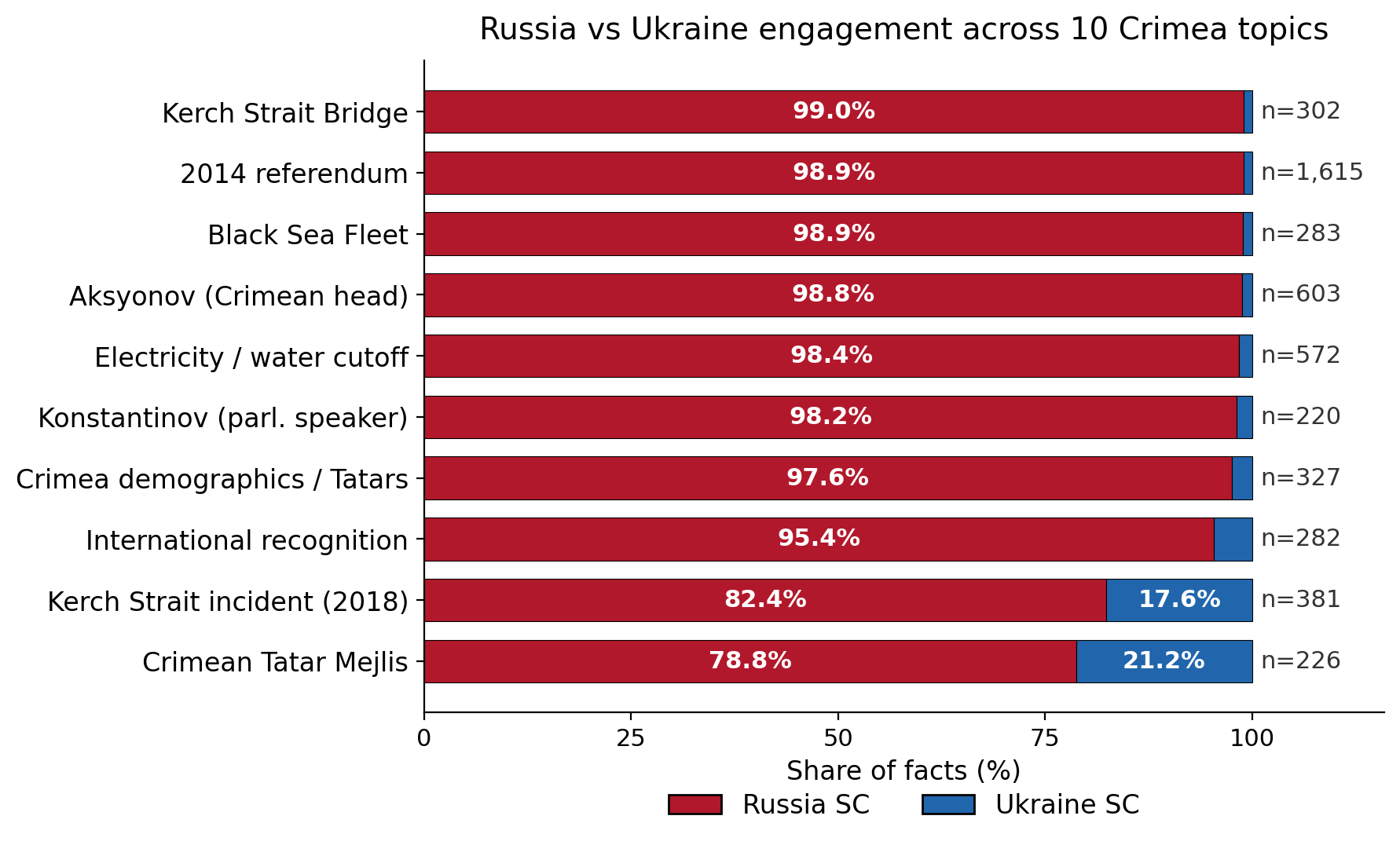}
        \caption{English-language state media (10 topics).}
        \label{fig:crimea-asymmetry}
    \end{subfigure}

    \vspace{0.5cm}

    \begin{subfigure}[t]{0.49\textwidth}
        \centering
        \includegraphics[width=\linewidth]{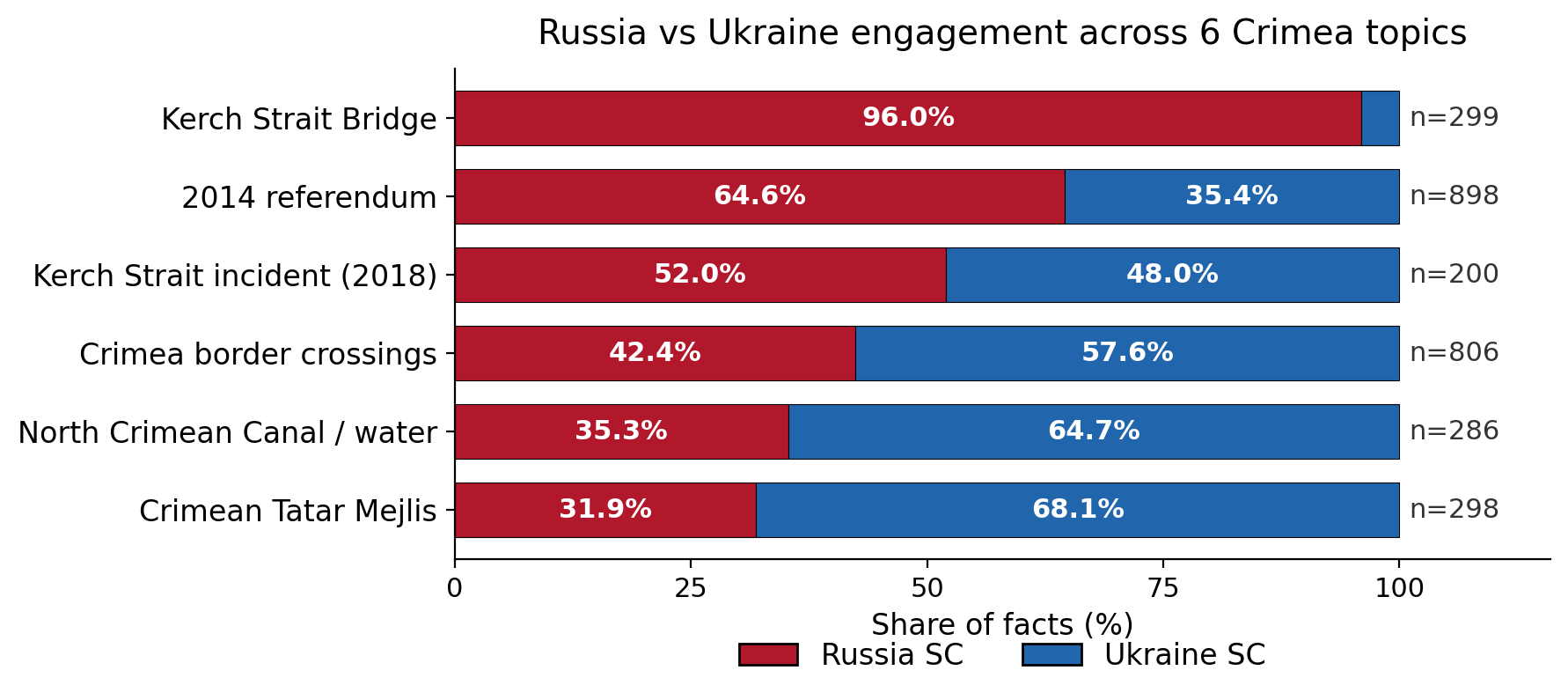}
        \caption{Native-language state media (6 topics).}
        \label{fig:crimea-asymmetry-domestic}
    \end{subfigure}

    \caption{Share of extracted facts from Russian and Ukrainian \smstate\ media across Crimea-related topics. $n$ indicates the number of facts in each topic.}
    \label{fig:crimea}
\end{figure}

\section{Case Study: State Media of Russia and Ukraine}
\label{sec:case-study}

While Section~\ref{sec:analysis} measured the prevalence of state media across the scale of LLM web data, we now zoom in, and perform a targeted case study on the ongoing conflict between Russia and Ukraine. In addition to the nations' territorial war, their respective state media have actively presented competing perspectives to domestic and international audiences through English-language and native-language (Russian, Ukrainian) articles respectively. We utilize the decontextualized facts approach from ~\citet{choi-etal-2021-decontextualization} to extract facts which can be understood without a source article. We do so from 269{,}001 unique articles (246{,}984 Russian and 22{,}017 Ukrainian; More details in Appendix~\ref{sec:fact_extraction}). 

\subsection{Topic Coverage Across State Media}
\label{sec:case-study-macro}

\paragraph{Methodology}
We embed the extracted facts and cluster them into 1{,}476 English topics and 247 native topics. For each topic, we compute a Laplace-smoothed log-odds ratio of Ukrainian vs. Russian representation to identify differences in coverage (corpus statistics, topic-skew distributions, and word clouds are in Appendix~\ref{sec:appendix_pipeline}).

\paragraph{English-language media}
Russian \smstate\ publishes substantially more English-language articles than Ukrainian \smstate\ (215{,}984 vs.\ 4{,}066). More than 1{,}100 of the 1{,}476 topics contain only Russian facts. Russia-skewed topics cover international affairs, such as Edward Snowden's asylum and the Libyan civil war. Ukraine-skewed topics focus on Ukrainian figures and issues, such as Shmyhal, Donbas, and the Holodomor. Only 92 topics contain at least 10 facts from each side before February 2022, and 55 after. These differences suggest that Russia's English-language \smstate\ more broadly covers international news, while Ukraine's English-language \smstate\ focuses more on Ukraine. 


\paragraph{Russian- and Ukrainian-language media}
The native-language corpora are more comparable in size: 31{,}000 Russian articles containing 432{,}329 facts and 17{,}951 Ukrainian articles containing 428{,}557 facts. Of the 247 topics, 236 contain facts from both sides, and 197 contain at least 10 facts from each. Both cover governance, economics, and international affairs. Russia-skewed topics focus on Russian figures and institutions, including Putin, Sobyanin, the Bank of Russia, and BRICS. Ukraine-skewed topics focus on public administration, including Ukrposhta, the NKREKP energy regulator, and Prime Ministers Groysman and Shmyhal. Unlike the English-language corpora, native-language coverage overlaps substantially.

\subsection{The Crimean Peninsula}

We next focus on the geopolitical dispute over the Crimean Peninsula between Russia and Ukraine to examine how their state media present different narratives about the same contested territory.

\subsubsection{Differences in Topic Coverage}

\begin{figure*}[t!]
  \centering
  \includegraphics[width=\textwidth]{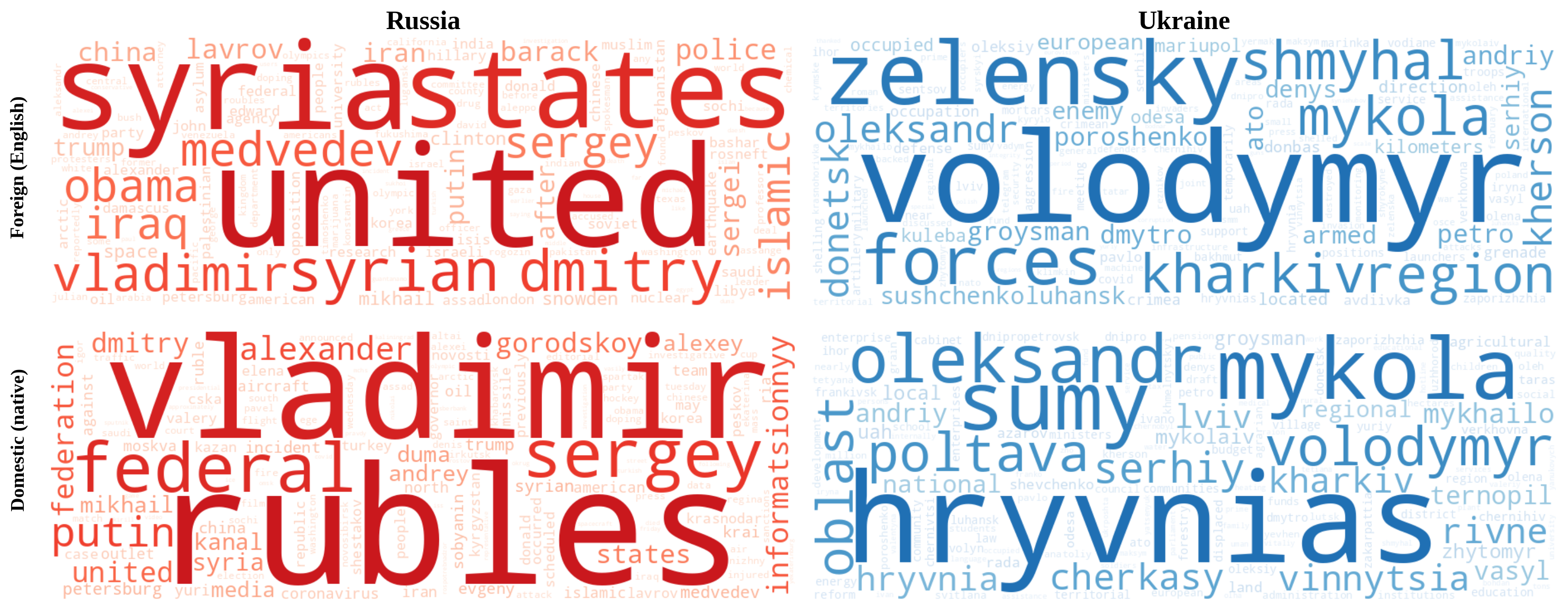}
  \caption{A 2-by-2 grid of word clouds for the corpora from Section~\ref{sec:case-study}. Rows indicate English-language or native-language, while columns indicate from Russia or Ukraine sources. Word size is proportional to weighted log-odds; larger words indicate more distinctive vocabulary. For readability, native words are translated to English.}
  \label{fig:wordcloud-corpus}
\end{figure*}

Figure~\ref{fig:wordcloud-corpus} shows word clouds of the distinctive vocabulary used by Russian and Ukraine state media. Notably, within a country, word clouds differ between foreign and domestic corpora.

\paragraph{English-language media}
Ten Crimea-related topics contain 4{,}811 facts. Russian \smstate\ accounts for 95--99\% of facts in most topics, including the 2014 referendum, Crimean leader Sergey Aksyonov, the Kerch Strait Bridge, and the Black Sea Fleet. Ukrainian \smstate\ has a larger presence in two topics: the 2018 Kerch Strait naval incident (17.6\% Ukrainian share) and the Crimean Tatar Mejlis (21.2\%; Figure~\ref{fig:crimea-asymmetry}). Because the Russian English-language corpus is approximately 50 times larger, these shares largely reflect the overall difference in corpus size. The exceptions indicate topics that receive relatively greater attention from Ukrainian outlets. Where coverage overlaps, the two sides sometimes describe the same events differently. For example, Russian \smstate\ states that ``the Crimea peninsula rejoined Russia,'' while Ukrainian \smstate\ states that ``Russia annexed Ukraine's Crimea peninsula'' (Appendix Tables~\ref{tab:example-facts} and~\ref{tab:example-facts-domestic}).

\paragraph{Russian- and Ukrainian-language media}
Six native-language Crimea topics contain 2{,}787 facts, with more balanced representation across the two sides (Figure~\ref{fig:crimea-asymmetry-domestic}). Russian \smstate\ dominates coverage of the Kerch Strait Bridge (96.0\% Russian share) and leads on the 2014 referendum (64.6\%). Ukrainian \smstate\ leads on the Mejlis (68.1\%) and the North Crimean Canal water dispute (64.7\%). Coverage of the Kerch Strait incident is nearly balanced (48.0\% Ukrainian share). Appendix~\ref{sec:crimea_llm_analysis} provides LLM-assisted analyses of framing in both language splits (method description in Appendix~\ref{sec:topic_framing_analysis}).

\subsubsection{Narrative Patterns}
\label{sec:ip_method}

\paragraph{Methodology}
To identify distinguishing narrative patterns between the two sides, we apply Information Pursuit (IP)~\citep{chan2025conformal}, an interpretable classification method (exact algorithmic details in Appendix~\ref{sec:appe_information_pursuit}). We label English-language Crimea articles by outlet country and use IP to distinguish Russian from Ukrainian reporting. IP asks binary questions about each article, such as ``Does the article describe Crimea as part of Russia?'' and an LLM answers the question. At each step, IP selects the question that minimizes the expected entropy of the label posterior. The query bank contains 19 questions about territorial status, contested terminology, and portrayals of key figures.

\paragraph{Findings}
The two most frequently selected questions concern whether Crimea is described as part of Russia (Q2) and whether the article focuses on diplomatic negotiations (Q12). IP selects these questions for 93.9\% and 85.7\% of sampled articles, respectively (Appendix Figure~\ref{fig:ip_queries}). These results identify territorial status and diplomatic framing as prominent signals used to distinguish Russian and Ukrainian reporting.
\begin{figure*}[t!]
    \centering
    \includegraphics[width=0.85\linewidth]{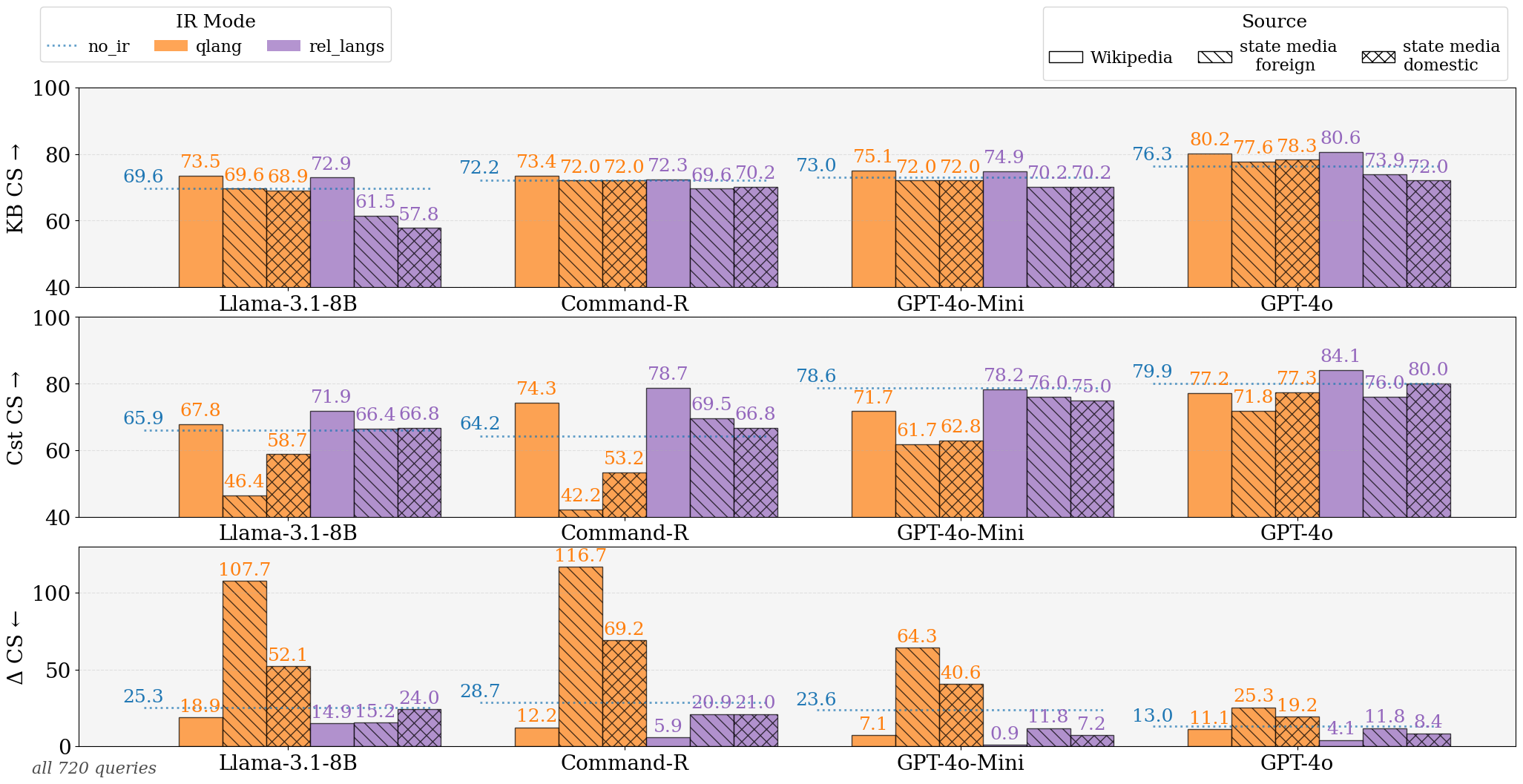}
    \caption[Concurrence score (CS) metrics over different retrieval settings, which vary the sources and the perspectives.]{Concurrence score (CS) metrics with retrieval varying across \textit{sources}: Wikipedia from~\cite{li-etal-2025-multilingual}, English (foreign) state media, and multilingual (domestic) state media; and across \textit{perspectives}: topic-level (\rellangs) vs. query-level (\qlang). Foreign state media at the query level yields the most divergent responses across languages, and topic-level retrieval is less divergent than query-level.}
    \label{fig:sm_cs_results}
\end{figure*}

\section{Retrieval over State Media}
\label{sec:rag}

The prior section compared state media narratives for two countries. However, the size and organized nature of \dataset\ unlocks the ability to study differences between nations' geopolitical narratives at scale. In this section, we showcase this utility through an experiment on how varying the state media database composition for retrieval-augmented generation (RAG) affects responses to controversial geopolitical queries. We find that RAG over state media actively decreases response factuality, unlike RAG over Wikipedia. Furthermore, retrieving from foreign-targeted state media results in greater divergence in model responses across different languages.

\subsection{Experimental Setup}

\paragraph{Task and metrics}
We use the \borderlines\ task of \citet{li2024land}, which consists of 720 queries on 251 disputed territories. Given a set of cross-lingual queries for a territory (e.g., asking about Crimea in Russian, Ukrainian, and English), the LLM's task is to choose which claimant country the territory belongs to. Responses are evaluated with three accuracy-based concurrence score (CS) metrics (details and formulas in Appendix~\ref{sec:cs_metrics}):
\begin{itemize}
    \item \kbcs\ $\uparrow$ measures \textit{factuality}, as agreement of the English response with the knowledge base (KB) of Wikipedia.
    \item  \cstcs\ $\uparrow$ measures cross-lingual \textit{consistency}, as pairwise agreement among a query's responses in different languages.
    \item \deltacs\ $\downarrow$ measures \textit{geopolitical bias}, as the relative difference in KB agreement between the controlling country's language and the other claimants' languages.
\end{itemize}

\citet{li-etal-2025-multilingual} extend the territorial disputes task by allowing an LLM to inform its answers by retrieving from Wikipedia. They experiment with the composition of the databases, and find that multilingual retrieval allows for the most cross-lingually consistent responses, while single-language retrieval leads to amplified geopolitical bias.

\paragraph{Retrieval settings}
Following the setup of~\citet{li-etal-2025-multilingual}, we apply OpenAI large embeddings to all documents in \dataset\, and retrieve those that are closest to each \borderlines\ query by cosine similarity.

While Wikipedia articles on the same topic have different authors, they are nonetheless subject to the same strict editorial standards and neutrality policy. Wikipedia can thus only be considered a single source. In this work, we study state media, which is an ideal test bed for studying real-world multilingual information as written by entities with divergent perspectives and interests, and consumed by audiences across the world.

Our RAG study thus adds the dimension of varied sources, to the existing dimension of varied perspectives, as illustrated in Appendix Figure~\ref{fig:sm_rag}.
The \textit{source} considers whether documents come from the English (foreign-targeted) split or the multilingual (domestic-targeted) split.\footnote{However, English articles are included in the domestic-targeted split for English-speaking countries.} The \textit{perspective} is the language of the query, and can be either at \textbf{query-level} or \textbf{topic-level}. Query-level means to retrieve only from documents of the country associated with its language (e.g., Russian outlets for the Russian query). Topic-level means to retrieve from the pooled state media of all claimants. Query-level retrieval approximates a language whose sources are dominated by one state's media, as with several languages in \S\ref{sec:analysis_lang}. Topic-level retrieval approximates a pool that mixes competing claims.

Again following the setup of~\citet{li-etal-2025-multilingual}, we evaluate four LLMs: GPT-4o, GPT-4o-mini~\cite{gpt4}, Command-R 35B~\cite{commandr}, and Llama 3.1 8B~\cite{llama3}. We include baselines with no retrieval (\noir) and Wikipedia retrieval.


\subsection{Results}

Results are shown in Figure~\ref{fig:sm_cs_results}. Our findings are:
\paragraph{Foreign-targeted retrieval causes more divergent responses}
\deltacs\ for foreign-targeted retrieval at the query-level (\qlang) is drastically amplified -- over 100\% for Llama and Command-R.  Recall that foreign-targeted means English-language articles; i.e., a RAG system queried in Arabic/Spanish/some language but reading that nation's English articles becomes more polarized in responses on average, than reading that language's native articles. This is likely a two-way street: both that the foreign-targeted English content is more polarizing, and RAG systems more readily adopt biased perspectives in English. Quantifying the relative strength of each effect is left to future work.

\paragraph{State media retrieval decreases factuality vs.\ Wikipedia}
While retrieval over Wikipedia increases \kbcs\ in 7 of 8 settings (1 neutral), retrieval over state media has a negative effect on \kbcs\ in 8 of 16 settings (6 neutral, 2 increases). Topic-level perspectives lead to the largest drops, particularly with domestic (multilingual) sources. These results underscore the real-world limitations of evaluating retrieval settings exclusively on Wikipedia. When we expose RAG systems to the diverse, polarized perspectives of state media, response factuality deteriorates. This may be because the diverse opinions and justifications to draw their immediate response away from that of the KB (Wikipedia).

\paragraph{Topic-level retrieval is more robust than query-level}
Comparing the topic-level (purple) and query-level (orange) bars, the former yields higher \cstcs\ and lower \deltacs\ for every model and source. This shows that RAG systems, when presented with differing narratives from different countries, can indeed make decisions that tend towards more consistency across languages. This robustness does not extend to factuality, however, as topic-level retrieval also has the lowest \kbcs.


\section{Conclusion}

State media outlets inherently project the authority of their home governments, and their prevalence and accessibility on the Internet means that their narratives increasingly pervade AI systems' training and retrieval corpora. To facilitate research into the influence of the global ecosystem of state media on the behavior of these systems, we introduced \dataset, a global-scale, open dataset of 118 million news articles from state media outlets from 136 countries in 114 languages.

Our analysis of \dataset\ revealed that the prevalence of news originating from editorially controlled has surged vs. from independent state media in English web crawls over the past decade, and also that editorially-controlled news was over-represented in web crawl data for low-resource languages such as Belarusian and Uyghur. Our case studies on Russia and Ukraine showed that foreign-targeted state media strategically prioritized contested international narratives over domestic affairs. Furthermore, we demonstrated that AI systems which retrieved over these state media sources -- especially foreign-targeted ones -- generated responses with lower factuality and higher cross-lingual inconsistency.

We release \dataset\ to enable the research community to critically analyze state media narratives and their downstream effects on AI systems across countries, languages, and audiences. We hope our research encourages development of dataset curation practices that account for editorial control and geopolitical bias. Ultimately, our work contributes toward the development of AI systems that are more robust to state-sponsored narratives and better equipped to navigate diverse geopolitical perspectives across languages and cultural contexts, and provide users with better informed responses.

\section*{Limitations}
We emphasize that state media comes in many forms and levels of control, and that our work does not seek to make judgments about the veracity of different outlets. We list several limitations below.
\begin{itemize}
    \item \textbf{Crawl coverage.} \dataset\ inherits the coverage of Common Crawl and the filtering of FineWeb and FineWeb-2. Page counts reflect accessibility and crawl policy, and not only the output or reach of outlets; countries whose state media is mainly radio and TV with little web presence, such as India, are underrepresented or absent. Temporal trends may be affected by changes in crawl composition, including publishers' restrictions on web crawlers. 
    \item \textbf{Static annotations.} State control labels come from the 2025 version of State Media Monitor and are applied to all years, although an outlet's level of state control may change over time.\footnote{The State Media Monitor website provides each outlet's typology for the three prior years. In follow-up work, we plan to obtain the full-time series from the project's maintainers and apply historical labels to earlier crawls.} Fixed labels mean that temporal trends reflect changes in page counts rather than reclassification, but they describe each outlet's status as of 2025 (\S\ref{sec:analysis_time}). Our URL list reflects current domains, so outlets that have changed domains may be under-represented in earlier crawls.
    \item \textbf{Prevalence.} The absolute proportion of state media in the English FineWeb data is low, at 0.2--0.6\% per crawl. However, because state media often lacks paywalls and is considered authoritative, it may exert an outsized influence on model responses. A more appropriate denominator would consider only news content, which we do not calculate here.
    \item \textbf{Scope of state media.} State media is only a fraction of each country's media landscape. In nations such as the US, the mainstream media is largely dominated by private outlets, whose biases and societal effects are well studied~\citep{groseclose2005measure,chang2023political,wang2025media}.
    \item \textbf{Model selection.} We follow the model settings of \citet{li-etal-2025-multilingual} to facilitate comparison with prior work. We leave to follow-up work to re-run the RAG experiments with more recent LLMs and IR systems, and verify the results still hold.
    
\end{itemize}

\section*{Ethics Statement}

\dataset\ contains state media content, including content from outlets that may disseminate propaganda or disinformation. We release it to support research on such narratives and their effects on AI systems, not to endorse any outlet or narrative. We label outlets according to the typology of State Media Monitor and do not make our own judgments about the veracity of individual outlets or articles. \dataset\ is derived from FineWeb and FineWeb-2, and is released under the same license (ODC-By), subject to the Common Crawl terms of use. We used AI tools to polish the writing; all text and figures were manually reviewed.


\bibliography{custom}

\appendix
\section{Details for \dataset}
\label{sec:dataset_details}

\subsection{Prompt for Obtaining URLs}
\label{sec:prompt_urls}

Figure~\ref{fig:sm_prompt1} shows the prompt used to generate URLs with GPT-3.5 Turbo. An entire row from State Media Monitor is given as input, with company, country, and list of assets. SMM's per-outlet reports sometimes mention links, but not in a structured or consistent form, which is why we generate candidate URLs. As described in Section~\ref{sec:collection}, the generated URLs are then manually verified, and URLs that fail to load are replaced via web search where possible.

\begin{figure*}[ht]
    \centering
    \begin{tcolorbox}[enhanced, width=\linewidth, boxrule=0.8mm] 
        \footnotesize
        \begin{verbatim}
You are provided a state-run media company, its country, and a list of its assets. For each asset, you
should provide the URL of the official website. Following the provided examples, output a JSON
dictionary, where each key is the asset name, and the value is the URL. If the URL is not found, use
the string 'No URL found.'

Company: U.S. Agency for Global Media (USAGM)
Country: United States
Assets: Radio: Voice of America, Radio Free Europe/Radio Liberty, Radio Free Asia, Office for Cuba 
Broadcasting, Middle East Broadcasting Networks
{"Voice of America": "https://www.voanews.com/",
"Radio Free Europe/Radio Liberty": "https://www.rferl.org/",
"Radio Free Asia": "https://www.rfa.org/english/",
"Office for Cuba Broadcasting": "https://www.martinoticias.com/",
"Middle East Broadcasting Networks": "https://www.alhurra.com/"}
[3 other exemplars; not shown]

Company: {company}
Country: {country}
Assets: {assets}
\end{verbatim}
    \end{tcolorbox}
    \centering
    \caption{Prompt for obtaining URLs for the assets of a state media company.}
    \label{fig:sm_prompt1}
\end{figure*}


\subsection{Detailed statistics for \dataset}
\label{sec:appendix_tables}

Table~\ref{tab:sample_articles} shows sample entries from each split. Table~\ref{tab:fwmulti_lang_pages} shows for each language in the multilingual split the proportions of \smindep, \smstate, total number of pages, and the proportions of state media pages out of the total. Table~\ref{tab:fwen_date_pages} shows these statistics for each temporal Common Crawl dump in the English split. Table~\ref{tab:fw_stats} shows these statistics for each country, grouped by split.

The 136 countries and 114 languages are obtained by filtering to countries with at least 10k pages and languages with at least 100k pages. Without these filters, the corpus covers 171 countries. 

\subsection{Corpus-level vs dump-level article deduplication}
\label{sec:app_dedup}
Fineweb performs deduplication of articles within Common Crawl dumps, instead of across the entire conjoined corpus. As explained in their blog post,\footnote{\url{https://huggingface.co/spaces/HuggingFaceFW/blogpost-fineweb-v1}}  language models trained on corpus-level deduplicated data had worse performance than those trained on dump-level. Our dataset \dataset\ inherits this level of deduplication. As a resource for auditing LLM training corpora, we find this to be reasonable. However, \dataset\ can also be used for retrieval, such as in this paper's \S\ref{sec:case-study} and \S\ref{sec:rag}.

In such cases, corpus-level deduplication should be utilized. We do though through the simple process of selecting unique URLs across the corpora, and keeping only the latest crawled version of a page. This results in 58m unique URLs and corpus-level deduplicated articles in this version of our dataset. \dataseten\ goes from 79m to 21m articles, while \datasetmulti\ goes from 39m to 37m articles. 

While this is a large drop, it is reasonable for Common Crawl to rescrape webpages, given that many change greatly over time. It is the case that state media article pages tend to stay more constant\footnote{We analyzed 3000 articles with duplicated URLs. 59\% were completely identical. The other 41\% had small differences in headers, footers, or page layout.}. As stated above, Fineweb developers explained their rationale for dump-level over corpus-level. And in our work, our rationale is that \dataset\ is a 118m corpora when viewed as a lens into LLM training corpora, while 58m when used as a retrieval corpora.


\begin{table*}[t!]
    \footnotesize
    \centering
    \setlength{\tabcolsep}{3pt}
    
\begin{tabularx}{\textwidth}{cccccccc}
    \toprule
    \textbf{Split} & \textbf{URL} & \textbf{Date} & \textbf{Language} & \textbf{Country} & \textbf{Subregion} & \textbf{Asset} & \textbf{Typology} \\
    \midrule
    2017-30 & azertag.az & 2017-07-22 & azj\_Latn & Azerbaijan & Middle East & Azertac & \smsc  \\
    \hfill \textbf{Text} & \multicolumn{7}{p{0.85\textwidth}}{Bakı, 8 iyul, Azərtac Təbii bal arılar tərəfindən çiçəklərdən və meyvə tumurcuqlarından alınaraq yığılan nektarlardan formalaşmış əvəzsiz qida məhsuludur. Mənşəyinə görə o çiçək və şirə balları kimi növlərə...} \\
    \midrule
    2017-34 & mediapermata.com.bn & 2017-08-16 & zsm\_Latn & Brunei & Southeast Asia & Media Permata & \smsc \\
    \hfill \textbf{Text} & \multicolumn{7}{p{0.85\textwidth}}{Hanoi, Vietnam mengekalkan pengharaman menangkap ikan lepas pantai di empat wilayah tengah setahun selepas kilang keluli milik Taiwan mengeluarkan toksik ke laut dan mengakibatkan bencana persekitaran paling teruk di negara itu...} \\
    \midrule
    2013-20 & thepeninsulaqatar.com & 2013-05-18 & en\_Latn & Qatar & Middle East & The Peninsula & \smca \\
    \hfill \textbf{Text} & \multicolumn{7}{p{0.85\textwidth}}{- Special Pages London: Manchester United, Manchester City, Arsenal and holders Chelsea all avoided each other in yesterday’s FA Cup fourth-round draw at Wembley, but only Chelsea and City are assured of a future in the competition...} \\
    \midrule
    2024-51 & iol.co.za & 2024-12-09 & en\_Latn & South Africa & Southern Africa & IOL & \smca \\
    \hfill \textbf{Text} & \multicolumn{7}{p{0.85\textwidth}}{On Wednesday, March 29, Nigerian Afrobeats star Davido ramped up the hype and excitement around his upcoming fourth studio album, “Timeless”, with the announcement of the project’s official track list. The album drops on Friday, March 31...} \\
    \bottomrule
\end{tabularx}

    \caption{Sample entries from \dataset, two from each split. We show the article text along with the metadata of split, URL, data, language, country, region, asset, and typology.}
    \label{tab:sample_articles}
\end{table*}

\begin{table}[t!]
    \scriptsize
    \centering
    \setlength{\tabcolsep}{3pt}
    \begin{tabularx}{\linewidth}{X r L c c | X r L c c}
\toprule
Lang & Hit \% & Total Pages & \smindep & \smstate & Lang & Hit \% & Total Pages & \smindep & \smstate \\
\midrule
rus & 1.0 & 699m & 4.2 & 94.0 & afr & 0.1 & 1.99m & 0.0 & 100.0 \\
cmn & 0.2 & 636m & 24.7 & 74.8 & tel & 0.0 & 1.96m & 5.1 & 94.9 \\
deu & 0.4 & 496m & 93.6 & 6.3 & hye & 0.6 & 1.76m & 0.2 & 99.8 \\
spa & 0.5 & 441m & 17.2 & 82.4 & khk & 4.0 & 1.62m & 0.0 & 100.0 \\
jpn & 0.0 & 400m & 89.8 & 10.2 & khm & 6.8 & 1.59m & 14.9 & 85.1 \\
fra & 0.8 & 360m & 86.5 & 13.2 & eus & 1.5 & 1.57m & 99.9 & 0.1 \\
ita & 0.1 & 239m & 59.4 & 40.6 & mya & 1.4 & 1.56m & 11.5 & 88.5 \\
por & 0.2 & 200m & 49.1 & 50.6 & lat & 0.0 & 1.47m & 78.5 & 18.7 \\
pol & 1.5 & 152m & 99.1 & 0.1 & uzn\_cyr & 1.8 & 1.36m & 1.1 & 80.4 \\
nld & 0.2 & 147m & 99.6 & 0.0 & uzn\_lat & 1.6 & 1.23m & 0.9 & 99.1 \\
ind & 0.1 & 100m & 94.9 & 5.1 & nno & 7.7 & 1.21m & 100.0 & 0.0 \\
tur & 3.4 & 95.1m & 1.6 & 91.4 & swh & 4.7 & 1.21m & 92.5 & 7.5 \\
ces & 0.6 & 66.1m & 94.5 & 5.3 & sin & 1.2 & 1.19m & 10.6 & 89.5 \\
arb & 5.3 & 62m & 9.0 & 83.3 & som & 1.9 & 1.07m & 36.4 & 63.6 \\
vie & 2.0 & 61.1m & 2.2 & 97.8 & kir & 1.9 & 1.07m & 0.1 & 99.9 \\
kor & 0.1 & 60.9m & 57.4 & 42.6 & pan & 0.1 & 944k & 100.0 & 0.0 \\
swe & 2.6 & 59.5m & 99.9 & 0.0 & arz & 4.5 & 853k & 0.8 & 98.7 \\
fas & 0.9 & 58.8m & 21.8 & 78.1 & cym & 3.8 & 832k & 100.0 & 0.0 \\
ron & 0.7 & 58.3m & 9.8 & 89.8 & nrm & 0.2 & 823k & 81.6 & 18.3 \\
ukr & 0.8 & 53.1m & 83.5 & 15.8 & tgk & 0.0 & 688k & 81.7 & 18.4 \\
hun & 3.1 & 49.9m & 0.1 & 99.8 & gle & 1.6 & 647k & 100.0 & 0.0 \\
ell & 0.1 & 47.4m & 37.7 & 61.0 & pbt & 5.9 & 640k & 68.8 & 31.2 \\
dan & 0.5 & 45.4m & 99.6 & 0.0 & srp\_lat & 7.0 & 586k & 4.2 & 95.5 \\
nob & 1.6 & 38.1m & 99.8 & 0.0 & hau & 7.0 & 570k & 97.4 & 2.7 \\
fin & 1.5 & 36.7m & 99.9 & 0.0 & ckb & 1.4 & 555k & 55.9 & 44.1 \\
tha & 0.2 & 35.9m & 4.2 & 95.8 & mlt & 4.5 & 489k & 0.4 & 99.6 \\
slk & 0.1 & 30m & 10.5 & 0.5 & kmr & 1.4 & 445k & 74.1 & 25.9 \\
bul & 1.1 & 26m & 25.3 & 74.4 & bew & 0.0 & 443k & 100.0 & 0.0 \\
hin & 0.2 & 22.1m & 89.3 & 10.8 & amh & 6.5 & 428k & 56.3 & 43.7 \\
bos & 4.8 & 21.2m & 6.6 & 93.1 & lao & 1.7 & 360k & 5.7 & 94.3 \\
cat & 3.1 & 17.1m & 0.0 & 84.6 & ltz & 4.1 & 355k & 100.0 & 0.0 \\
ben & 0.1 & 15.2m & 63.7 & 36.3 & fry & 0.0 & 350k & 100.0 & 0.0 \\
heb & 0.1 & 14.5m & 53.2 & 42.2 & div & 1.7 & 349k & 0.0 & 100.0 \\
lit & 1.8 & 13.5m & 100.0 & 0.0 & epo & 0.8 & 336k & 0.0 & 100.0 \\
slv & 2.7 & 12.1m & 99.4 & 0.3 & kin & 0.0 & 326k & 100.0 & 0.0 \\
ekk & 0.8 & 10.2m & 100.0 & 0.0 & yue & 0.2 & 315k & 0.0 & 100.0 \\
zsm & 1.7 & 9.42m & 0.0 & 100.0 & ars & 0.3 & 298k & 4.5 & 91.1 \\
als & 1.1 & 8.6m & 27.5 & 49.6 & fao & 0.0 & 291k & 100.0 & 0.0 \\
lvs & 0.1 & 8.03m & 0.0 & 0.0 & snd & 0.0 & 258k & 0.0 & 100.0 \\
azj & 2.5 & 7.29m & 0.0 & 100.0 & xho & 0.0 & 254k & 0.0 & 100.0 \\
hrv & 1.2 & 6.2m & 14.7 & 85.2 & tuk & 8.3 & 238k & 0.0 & 62.8 \\
tam & 0.6 & 5.53m & 3.5 & 96.5 & hat & 0.0 & 224k & 100.0 & 0.0 \\
npi & 0.4 & 4.89m & 3.5 & 78.0 & gla & 6.4 & 222k & 100.0 & 0.0 \\
urd & 1.7 & 4.81m & 44.5 & 55.5 & ceb & 0.1 & 205k & 0.0 & 100.0 \\
mkd & 1.4 & 4.15m & 74.0 & 26.0 & bak & 0.0 & 183k & 0.0 & 100.0 \\
srp\_cyr & 5.1 & 4.15m & 0.2 & 99.9 & mri & 6.8 & 167k & 100.0 & 0.0 \\
mar & 0.0 & 3.91m & 16.7 & 83.3 & uig & 17.8 & 166k & 0.0 & 58.5 \\
kat & 3.3 & 3.71m & 0.0 & 100.0 & tat & 0.0 & 161k & 0.0 & 100.0 \\
kaz & 5.3 & 3.34m & 0.0 & 68.0 & bod & 4.4 & 161k & 2.1 & 97.9 \\
mal & 0.0 & 3.32m & 13.6 & 86.4 & mww & 0.2 & 159k & 100.0 & 0.0 \\
isl & 1.8 & 3.01m & 100.0 & 0.0 & hyw & 0.1 & 152k & 28.9 & 71.1 \\
glg & 0.0 & 2.52m & 0.0 & 21.1 & ydd & 0.0 & 135k & 100.0 & 0.0 \\
kan & 0.0 & 2.39m & 50.0 & 50.0 & zul & 4.5 & 127k & 0.0 & 100.0 \\
ary & 0.9 & 2.37m & 9.1 & 83.0 & sdh & 1.2 & 114k & 0.3 & 99.7 \\
fil & 0.1 & 2.35m & 61.7 & 38.3 & cos & 0.1 & 111k & 100.0 & 0.0 \\
guj & 0.0 & 2.13m & 41.6 & 58.4 & smo & 0.3 & 111k & 100.0 & 0.0 \\
bel & 7.8 & 2.1m & 1.6 & 59.9 & nya & 0.3 & 103k & 0.0 & 100.0 \\
\bottomrule
\end{tabularx}

    \caption{For each of the 114 languages from \datasetmulti, the hit \% from state media outlets over all pages in Fineweb-2, the total pages (all content types), and the proportions from independent (\smindep) and controlled (\smstate) state media. Percentages may not sum to 100 where some outlets lack a typology label.}
    \label{tab:fwmulti_lang_pages}
\end{table}

\begin{table}[t!]
    \scriptsize
    \centering
    \setlength{\tabcolsep}{2pt}
    \begin{tabularx}{\linewidth}{X r L c c | X r L c c}
\toprule
Split & Hit \% & Total Pages & \smindep & \smstate & Split & Hit \% & Total Pages & \smindep & \smstate \\
\midrule
2013-20 & 0.4 & 215m & 61.2 & 38.6 & 2019-13 & 0.2 & 219m & 33.5 & 64.3 \\
2013-48 & 0.4 & 220m & 72.9 & 26.7 & 2019-18 & 0.3 & 223m & 36.3 & 61.4 \\
2014-10 & 0.4 & 217m & 65.6 & 34.0 & 2019-22 & 0.3 & 227m & 35.1 & 63.2 \\
2014-15 & 0.4 & 200m & 63.2 & 36.3 & 2019-26 & 0.3 & 217m & 37.2 & 61.6 \\
2014-23 & 0.4 & 234m & 68.4 & 31.4 & 2019-30 & 0.3 & 219m & 36.3 & 62.1 \\
2014-35 & 0.4 & 216m & 64.8 & 34.9 & 2019-35 & 0.3 & 239m & 37.3 & 61.5 \\
2014-41 & 0.4 & 224m & 64.1 & 35.7 & 2019-39 & 0.3 & 206m & 41.0 & 57.7 \\
2014-42 & 0.4 & 201m & 58.0 & 41.6 & 2019-43 & 0.3 & 223m & 31.9 & 67.3 \\
2014-49 & 0.4 & 171m & 55.0 & 44.7 & 2019-47 & 0.2 & 208m & 37.3 & 61.1 \\
2014-52 & 0.5 & 212m & 63.7 & 36.1 & 2019-51 & 0.3 & 188m & 35.9 & 62.8 \\
2015-06 & 0.4 & 193m & 63.2 & 36.5 & 2020-05 & 0.2 & 253m & 32.1 & 66.3 \\
2015-11 & 0.5 & 195m & 61.3 & 38.5 & 2020-10 & 0.3 & 194m & 33.0 & 65.6 \\
2015-14 & 0.4 & 181m & 59.0 & 40.8 & 2020-16 & 0.2 & 238m & 31.7 & 66.5 \\
2015-18 & 0.5 & 210m & 63.8 & 36.0 & 2020-24 & 0.2 & 213m & 29.2 & 68.7 \\
2015-22 & 0.5 & 204m & 61.7 & 38.1 & 2020-29 & 0.2 & 256m & 29.9 & 66.9 \\
2015-27 & 0.4 & 180m & 62.3 & 37.4 & 2020-34 & 0.3 & 197m & 33.5 & 64.3 \\
2015-32 & 0.5 & 189m & 63.6 & 36.2 & 2020-40 & 0.2 & 293m & 29.2 & 67.6 \\
2015-35 & 0.5 & 191m & 63.1 & 36.7 & 2020-45 & 0.3 & 221m & 35.0 & 62.9 \\
2015-40 & 0.4 & 150m & 61.0 & 38.8 & 2020-50 & 0.3 & 221m & 33.2 & 64.9 \\
2015-48 & 0.5 & 187m & 63.6 & 36.2 & 2021-04 & 0.2 & 293m & 33.8 & 63.1 \\
2016-07 & 0.5 & 180m & 63.1 & 36.7 & 2021-10 & 0.2 & 236m & 35.8 & 62.2 \\
2016-18 & 0.3 & 157m & 57.3 & 42.3 & 2021-17 & 0.2 & 280m & 36.4 & 60.5 \\
2016-22 & 0.4 & 160m & 58.6 & 41.0 & 2021-21 & 0.2 & 228m & 37.8 & 60.1 \\
2016-26 & 0.4 & 132m & 63.9 & 35.7 & 2021-25 & 0.2 & 214m & 36.3 & 62.2 \\
2016-30 & 0.6 & 178m & 66.7 & 33.1 & 2021-31 & 0.2 & 304m & 37.9 & 60.6 \\
2016-36 & 0.5 & 172m & 63.7 & 36.0 & 2021-39 & 0.2 & 260m & 37.6 & 61.0 \\
2016-40 & 0.6 & 200m & 39.7 & 59.8 & 2021-43 & 0.2 & 301m & 35.9 & 61.0 \\
2016-44 & 0.5 & 266m & 39.6 & 58.5 & 2021-49 & 0.3 & 206m & 37.5 & 60.4 \\
2016-50 & 0.6 & 252m & 36.7 & 61.3 & 2022-05 & 0.2 & 260m & 37.5 & 60.1 \\
2017-04 & 0.6 & 267m & 34.0 & 64.3 & 2022-21 & 0.2 & 326m & 34.6 & 62.4 \\
2017-09 & 0.6 & 277m & 38.8 & 59.6 & 2022-27 & 0.2 & 287m & 37.2 & 59.8 \\
2017-13 & 0.5 & 320m & 33.7 & 64.6 & 2022-33 & 0.3 & 215m & 36.8 & 61.1 \\
2017-17 & 0.4 & 325m & 32.7 & 65.8 & 2022-40 & 0.2 & 295m & 30.2 & 68.3 \\
2017-22 & 0.5 & 244m & 33.6 & 64.8 & 2022-49 & 0.2 & 308m & 29.9 & 67.2 \\
2017-26 & 0.4 & 275m & 33.9 & 64.5 & 2023-06 & 0.2 & 307m & 29.7 & 67.9 \\
2017-30 & 0.5 & 241m & 28.6 & 70.3 & 2023-14 & 0.2 & 295m & 30.7 & 66.8 \\
2017-34 & 0.4 & 290m & 35.3 & 62.8 & 2023-23 & 0.2 & 306m & 30.4 & 66.4 \\
2017-39 & 0.4 & 248m & 35.9 & 62.8 & 2023-40 & 0.2 & 327m & 26.2 & 70.5 \\
2017-43 & 0.3 & 303m & 38.9 & 60.1 & 2023-50 & 0.2 & 318m & 26.8 & 69.8 \\
2017-47 & 0.4 & 253m & 38.6 & 61.0 & 2024-10 & 0.2 & 276m & 17.1 & 80.9 \\
2017-51 & 0.4 & 250m & 36.1 & 61.2 & 2024-18 & 0.2 & 247m & 16.1 & 82.3 \\
2018-05 & 0.4 & 266m & 31.4 & 66.5 & 2024-22 & 0.2 & 237m & 18.2 & 79.9 \\
2018-09 & 0.3 & 273m & 37.4 & 61.0 & 2024-26 & 0.2 & 237m & 19.5 & 78.1 \\
2018-13 & 0.4 & 263m & 32.9 & 65.0 & 2024-30 & 0.2 & 215m & 17.2 & 80.6 \\
2018-17 & 0.3 & 238m & 41.1 & 55.8 & 2024-33 & 0.2 & 191m & 17.4 & 80.6 \\
2018-22 & 0.3 & 220m & 34.1 & 60.4 & 2024-38 & 0.2 & 242m & 15.4 & 82.6 \\
2018-26 & 0.3 & 250m & 38.6 & 59.6 & 2024-42 & 0.2 & 206m & 15.4 & 82.6 \\
2018-30 & 0.3 & 277m & 32.8 & 64.4 & 2024-46 & 0.2 & 225m & 17.6 & 80.4 \\
2018-34 & 0.2 & 217m & 32.7 & 64.4 & 2024-51 & 0.2 & 179m & 15.7 & 82.3 \\
2018-39 & 0.4 & 232m & 35.6 & 62.1 & 2025-05 & 0.2 & 266m & 13.1 & 85.5 \\
2018-43 & 0.3 & 247m & 44.6 & 52.4 & 2025-08 & 0.2 & 224m & 13.4 & 85.1 \\
2018-47 & 0.3 & 221m & 35.0 & 63.8 & 2025-13 & 0.2 & 233m & 7.0 & 91.0 \\
2018-51 & 0.2 & 265m & 33.8 & 64.5 & 2025-18 & 0.2 & 241m & 5.8 & 92.5 \\
2019-04 & 0.3 & 234m & 34.3 & 62.7 & 2025-21 & 0.2 & 219m & 5.5 & 93.0 \\
2019-09 & 0.2 & 246m & 35.8 & 61.3 & 2025-26 & 0.2 & 197m & 5.7 & 93.1 \\
\bottomrule
\end{tabularx}

    \caption{For each Common Crawl dump of \dataseten, the hit \% from state media outlets over all pages in English Fineweb, the total pages (all content types), and the proportions from independent (\smindep) and controlled (\smstate) state media.}
    \label{tab:fwen_date_pages}
\end{table}

\begin{table*}[t!]
    \scriptsize
    \centering
    \setlength{\tabcolsep}{3pt}
    \begin{tabularx}{\textwidth}{X r l c c r l c c | X r l c c r l c c}
\toprule
 & \multicolumn{4}{c}{English} & \multicolumn{4}{c|}{Multilingual} & & \multicolumn{4}{c}{English} & \multicolumn{4}{c}{Multilingual} \\
\cmidrule(lr){2-5} \cmidrule(lr){6-9} \cmidrule(lr){11-14} \cmidrule(lr){15-18}
Country & Rank & Pages & \smindep & \smstate & Rank & Pages & \smindep & \smstate & Country & Rank & Pages & \smindep & \smstate & Rank & Pages & \smindep & \smstate \\
\midrule
United States & 1 & 15.7m & 41.7 & 58.3 & 36 & 208k & 0.4 & 99.6 & Myanmar & 69 & 32.5k & 0.0 & 100.0 & 120 & 578 & 0.0 & 100.0 \\
Russia & 2 & 8.26m & 0.0 & 100.0 & 1 & 6.22m & 0.0 & 100.0 & Angola & 70 & 30.6k & 0.0 & 100.0 & 51 & 84.8k & 0.0 & 100.0 \\
Australia & 3 & 6.67m & 100.0 & 0.0 & 72 & 35k & 100.0 & 0.0 & Palestine & 71 & 29.1k & 0.0 & 100.0 & 74 & 32.6k & 0.0 & 100.0 \\
United Kingdom & 4 & 6.35m & 100.0 & 0.0 & 53 & 81.3k & 100.0 & 0.0 & Nepal & 72 & 29k & 0.0 & 99.2 & 87 & 18.5k & 0.0 & 80.8 \\
Canada & 5 & 4.98m & 100.0 & 0.0 & 26 & 363k & 100.0 & 0.0 & Malawi & 73 & 28.8k & 0.0 & 100.0 & 123 & 262 & 0.0 & 100.0 \\
Singapore & 6 & 4.61m & 0.0 & 100.0 & 25 & 386k & 0.0 & 100.0 & Barbados & 74 & 25.8k & 0.0 & 100.0 & - & 0 & - & - \\
South Africa & 7 & 3.75m & 0.0 & 100.0 & 85 & 21.3k & 0.0 & 100.0 & Romania & 75 & 25.4k & 0.0 & 100.0 & 23 & 396k & 0.0 & 100.0 \\
Qatar & 8 & 3.58m & 0.0 & 100.0 & 14 & 803k & 0.0 & 100.0 & Netherlands & 76 & 24.8k & 100.0 & 0.0 & 38 & 180k & 100.0 & 0.0 \\
Ireland & 9 & 2.65m & 100.0 & 0.0 & 104 & 9,817 & 100.0 & 0.0 & Bahamas & 77 & 23.4k & 0.0 & 100.0 & - & 0 & - & - \\
China & 10 & 2.35m & 0.0 & 100.0 & 20 & 419k & 0.0 & 100.0 & Kazakhstan & 78 & 22.1k & 0.0 & 31.6 & 21 & 404k & 0.0 & 57.4 \\
Turkey & 11 & 2.3m & 0.0 & 100.0 & 2 & 3.42m & 0.0 & 93.5 & Kuwait & 79 & 21.2k & 0.0 & 100.0 & 109 & 7,609 & 0.0 & 100.0 \\
UAE & 12 & 2.02m & 0.0 & 97.9 & 13 & 833k & 5.7 & 62.3 & Georgia & 80 & 19.2k & 0.0 & 100.0 & 42 & 137k & 0.0 & 100.0 \\
Vietnam & 13 & 1.68m & 0.0 & 100.0 & 8 & 1.22m & 0.0 & 100.0 & Malta & 81 & 19k & 0.0 & 100.0 & 84 & 22k & 0.0 & 100.0 \\
France & 14 & 1.36m & 100.0 & 0.0 & 5 & 2.08m & 100.0 & 0.0 & Uzbekistan & 82 & 18.8k & 0.0 & 100.0 & 59 & 59.8k & 0.0 & 100.0 \\
Egypt & 15 & 1.09m & 0.0 & 100.0 & 30 & 296k & 0.0 & 100.0 & Moldova & 83 & 18.3k & 15.4 & 84.6 & 81 & 24.8k & 22.9 & 77.1 \\
Cambodia & 16 & 1.09m & 0.0 & 9.6 & 52 & 82k & 0.0 & 100.0 & Liberia & 84 & 17.1k & 0.0 & 100.0 & - & 0 & - & - \\
Germany & 17 & 1.02m & 100.0 & 0.0 & 3 & 2.49m & 100.0 & 0.0 & Hong Kong & 85 & 15.7k & 0.0 & 100.0 & 31 & 272k & 0.0 & 100.0 \\
New Zealand & 18 & 995k & 100.0 & 0.0 & 98 & 11.4k & 100.0 & 0.0 & Somalia & 86 & 14.5k & 0.0 & 100.0 & 93 & 13.5k & 0.0 & 100.0 \\
Iran & 19 & 664k & 0.0 & 100.0 & 17 & 496k & 0.0 & 100.0 & Tonga & 87 & 13.8k & 0.0 & 0.0 & 122 & 288 & 0.0 & 0.0 \\
Slovenia & 20 & 597k & 100.0 & 0.0 & 29 & 327k & 100.0 & 0.0 & Peru & 88 & 13.8k & 0.0 & 100.0 & 45 & 112k & 0.0 & 100.0 \\
Saudi Arabia & 21 & 532k & 0.0 & 100.0 & 19 & 437k & 0.0 & 100.0 & Bahrain & 89 & 10.8k & 0.0 & 100.0 & 114 & 3,286 & 0.0 & 100.0 \\
Zimbabwe & 22 & 508k & 0.0 & 100.0 & 125 & 12 & 0.0 & 100.0 & North Korea & 90 & 10.7k & 0.0 & 100.0 & 86 & 19.2k & 0.0 & 100.0 \\
Switzerland & 23 & 466k & 100.0 & 0.0 & 11 & 900k & 100.0 & 0.0 & Iraq & 91 & 9,667 & 0.0 & 100.0 & 65 & 52.3k & 0.0 & 100.0 \\
Ghana & 24 & 386k & 0.0 & 100.0 & - & 0 & - & - & Lithuania & 92 & 9,434 & 100.0 & 0.0 & 32 & 253k & 100.0 & 0.0 \\
Hungary & 25 & 293k & 0.0 & 100.0 & 7 & 1.53m & 0.0 & 100.0 & Brazil & 93 & 8,619 & 0.0 & 100.0 & 57 & 74.4k & 0.0 & 100.0 \\
Uganda & 26 & 293k & 0.0 & 100.0 & 127 & 1 & 0.0 & 100.0 & Lebanon & 94 & 8,187 & 0.0 & 100.0 & 107 & 9,062 & 0.0 & 100.0 \\
Nigeria & 27 & 282k & 0.0 & 100.0 & 121 & 323 & 0.0 & 100.0 & Turkmenistan & 95 & 7,567 & 0.0 & 100.0 & 106 & 9,157 & 0.0 & 100.0 \\
Brunei & 28 & 249k & 0.0 & 100.0 & 56 & 75.3k & 0.0 & 100.0 & Afghanistan & 96 & 7,239 & 0.0 & 100.0 & 89 & 18.3k & 0.0 & 100.0 \\
Bhutan & 29 & 244k & 80.4 & 19.6 & - & 0 & - & - & Croatia & 97 & 5,720 & 0.0 & 100.0 & 58 & 63.3k & 0.0 & 100.0 \\
Czechia & 30 & 238k & 4.5 & 95.5 & 22 & 402k & 94.7 & 5.3 & The Maldives & 98 & 5,524 & 0.0 & 100.0 & 111 & 6,092 & 0.0 & 100.0 \\
Kenya & 31 & 210k & 0.0 & 100.0 & 117 & 1,773 & 0.0 & 100.0 & Cape Verde & 99 & 5,161 & 0.0 & 100.0 & 80 & 26.5k & 0.0 & 100.0 \\
Indonesia & 32 & 208k & 7.5 & 92.5 & 66 & 50.1k & 100.0 & 0.0 & Burundi & 100 & 5,048 & 0.0 & 100.0 & 112 & 5,232 & 0.0 & 100.0 \\
Venezuela & 33 & 192k & 0.0 & 100.0 & 44 & 115k & 0.0 & 100.0 & Kyrgyzstan & 101 & 4,938 & 0.0 & 100.0 & 62 & 56.6k & 0.0 & 100.0 \\
Pakistan & 34 & 185k & 0.0 & 100.0 & 115 & 2,547 & 0.0 & 100.0 & Algeria & 102 & 4,783 & 0.0 & 100.0 & 46 & 97.6k & 0.0 & 100.0 \\
St Kitts \& Nevis & 35 & 183k & 0.0 & 100.0 & - & 0 & - & - & Thailand & 103 & 4,779 & 0.0 & 100.0 & 64 & 54.2k & 0.0 & 100.0 \\
Azerbaijan & 36 & 181k & 0.0 & 100.0 & 24 & 386k & 0.0 & 100.0 & Austria & 104 & 4,368 & 100.0 & 0.0 & 50 & 89k & 100.0 & 0.0 \\
Cuba & 37 & 168k & 0.0 & 100.0 & 35 & 217k & 0.0 & 100.0 & Kosovo & 105 & 2,884 & 0.0 & 0.0 & 82 & 24.5k & 0.0 & 0.0 \\
Namibia & 38 & 166k & 0.0 & 100.0 & 124 & 103 & 0.0 & 100.0 & Slovakia & 106 & 2,595 & 37.8 & 0.0 & 78 & 26.6k & 9.7 & 0.0 \\
Ukraine & 39 & 160k & 2.1 & 97.9 & 28 & 335k & 76.2 & 23.8 & Sudan & 107 & 2,386 & 0.0 & 100.0 & 105 & 9,532 & 0.0 & 100.0 \\
Spain & 40 & 122k & 14.5 & 85.5 & 9 & 1.18m & 17.6 & 75.4 & Bosnia & 108 & 2,146 & 0.0 & 100.0 & 49 & 92.8k & 0.0 & 100.0 \\
Fiji & 41 & 118k & 100.0 & 0.0 & - & 0 & - & - & Norway & 109 & 2,091 & 100.0 & 0.0 & 15 & 692k & 100.0 & 0.0 \\
Solomon Islands & 42 & 110k & 100.0 & 0.0 & - & 0 & - & - & Iceland & 110 & 2,090 & 100.0 & 0.0 & 63 & 54.5k & 100.0 & 0.0 \\
Poland & 43 & 109k & 78.1 & 0.0 & 4 & 2.34m & 95.0 & 0.0 & Denmark & 111 & 1,063 & 100.0 & 0.0 & 34 & 237k & 100.0 & 0.0 \\
Taiwan & 44 & 105k & 97.5 & 2.5 & 41 & 139k & 99.7 & 0.3 & Morocco & 112 & 952 & 0.0 & 100.0 & 69 & 42.8k & 0.0 & 100.0 \\
Oman & 45 & 100k & 0.0 & 100.0 & 110 & 6,443 & 0.0 & 100.0 & Greece & 113 & 932 & 0.0 & 100.0 & 75 & 28.6k & 0.0 & 100.0 \\
Seychelles & 46 & 93.3k & 0.0 & 0.0 & 116 & 2,427 & 0.0 & 0.0 & Paraguay & 114 & 397 & 0.0 & 100.0 & 97 & 12.3k & 0.0 & 100.0 \\
South Korea & 47 & 90.2k & 100.0 & 0.0 & 47 & 96.8k & 100.0 & 0.0 & Portugal & 115 & 287 & 100.0 & 0.0 & 60 & 57.1k & 100.0 & 0.0 \\
Sri Lanka & 48 & 90k & 0.0 & 100.0 & 94 & 13.3k & 0.0 & 100.0 & Italy & 116 & 193 & 0.0 & 100.0 & 43 & 128k & 0.0 & 100.0 \\
Zambia & 49 & 89k & 0.0 & 100.0 & - & 0 & - & - & Argentina & 117 & 124 & 0.0 & 100.0 & 76 & 27.8k & 0.0 & 100.0 \\
Jordan & 50 & 85.5k & 0.0 & 100.0 & 48 & 93.3k & 0.0 & 100.0 & Chile & 118 & 80 & 0.0 & 100.0 & 37 & 182k & 0.0 & 100.0 \\
Philippines & 51 & 82.2k & 0.0 & 100.0 & 118 & 946 & 0.0 & 100.0 & Estonia & 119 & 65 & 100.0 & 0.0 & 40 & 161k & 100.0 & 0.0 \\
Belarus & 52 & 76.6k & 0.0 & 100.0 & 12 & 852k & 0.0 & 100.0 & Tunisia & 121 & 60 & 0.0 & 100.0 & 67 & 46.3k & 0.0 & 100.0 \\
Serbia & 53 & 74.7k & 0.0 & 100.0 & 10 & 991k & 0.0 & 100.0 & Luxembourg & 120 & 60 & 100.0 & 0.0 & 92 & 14.7k & 100.0 & 0.0 \\
Belgium & 54 & 73.4k & 100.0 & 0.0 & 18 & 477k & 100.0 & 0.0 & Ecuador & 122 & 46 & 0.0 & 100.0 & 61 & 56.8k & 0.0 & 100.0 \\
Armenia & 55 & 69.2k & 0.0 & 100.0 & 71 & 37.4k & 0.0 & 100.0 & Nicaragua & 123 & 37 & 0.0 & 100.0 & 33 & 239k & 0.0 & 100.0 \\
Syria & 56 & 64.1k & 0.0 & 100.0 & 39 & 178k & 0.0 & 100.0 & Mozambique & 124 & 15 & 0.0 & 100.0 & 96 & 12.7k & 0.0 & 100.0 \\
Finland & 57 & 62.3k & 100.0 & 0.0 & 16 & 522k & 100.0 & 0.0 & Israel & 125 & 12 & 100.0 & 0.0 & 103 & 10k & 100.0 & 0.0 \\
Bangladesh & 58 & 60.1k & 0.0 & 100.0 & 113 & 3,595 & 0.0 & 100.0 & Mexico & 126 & 11 & 54.5 & 45.5 & 95 & 13.3k & 72.0 & 28.0 \\
Tanzania & 59 & 58.5k & 0.0 & 100.0 & 119 & 757 & 0.0 & 100.0 & Latvia & 127 & 10 & 0.0 & 0.0 & 99 & 11.3k & 0.0 & 0.0 \\
Sweden & 60 & 51.8k & 100.0 & 0.0 & 6 & 1.65m & 100.0 & 0.0 & Colombia & 128 & 7 & 0.0 & 100.0 & 68 & 43.4k & 7.3 & 92.7 \\
Eritrea & 61 & 45.5k & 0.0 & 100.0 & 126 & 8 & 0.0 & 100.0 & El Salvador & 129 & 6 & 0.0 & 100.0 & 77 & 27.3k & 0.0 & 100.0 \\
Bulgaria & 62 & 41.1k & 42.3 & 57.7 & 27 & 340k & 12.7 & 87.3 & Benin & 130 & 3 & 0.0 & 100.0 & 101 & 11.1k & 0.0 & 100.0 \\
Malaysia & 63 & 38k & 0.0 & 100.0 & 108 & 7,998 & 0.0 & 100.0 & Guatemala & 131 & 1 & 0.0 & 100.0 & 73 & 34.6k & 0.0 & 75.9 \\
Cameroon & 64 & 36.1k & 0.0 & 100.0 & 90 & 17.7k & 0.0 & 100.0 & Côte d'Ivoire & - & 0 & - & - & 79 & 26.6k & 71.3 & 28.7 \\
Yemen & 65 & 35.3k & 0.0 & 76.1 & 70 & 40.3k & 0.0 & 68.6 & Burkina Faso & - & 0 & - & - & 83 & 22.6k & 0.0 & 100.0 \\
Ethiopia & 66 & 34.7k & 0.0 & 100.0 & 91 & 17k & 0.0 & 100.0 & Andorra & - & 0 & - & - & 88 & 18.4k & 0.0 & 100.0 \\
Mongolia & 67 & 34.1k & 0.0 & 100.0 & 54 & 79.3k & 0.0 & 100.0 & Albania & - & 0 & - & - & 100 & 11.2k & 0.0 & 100.0 \\
Japan & 68 & 33.8k & 100.0 & 0.0 & 55 & 77.5k & 100.0 & 0.0 & Uruguay & - & 0 & - & - & 102 & 10.1k & 0.0 & 100.0 \\
\bottomrule
\end{tabularx}
    \caption{Dataset statistics for \dataseten\ and \datasetmulti\ over all 136 countries. For each country and split, we show the ordinal rank, the number of pages (state media only), and the percentage of articles from either independent (\smindep) or controlled (\smstate) state media. Percentages may not sum to 100 where some outlets lack a typology label.}
    \label{tab:fw_stats}
\end{table*}





\section{Details for the RAG Experiment}
\label{sec:rag_details}
\begin{figure*}[t!]
    \centering
    \includegraphics[width=\linewidth]{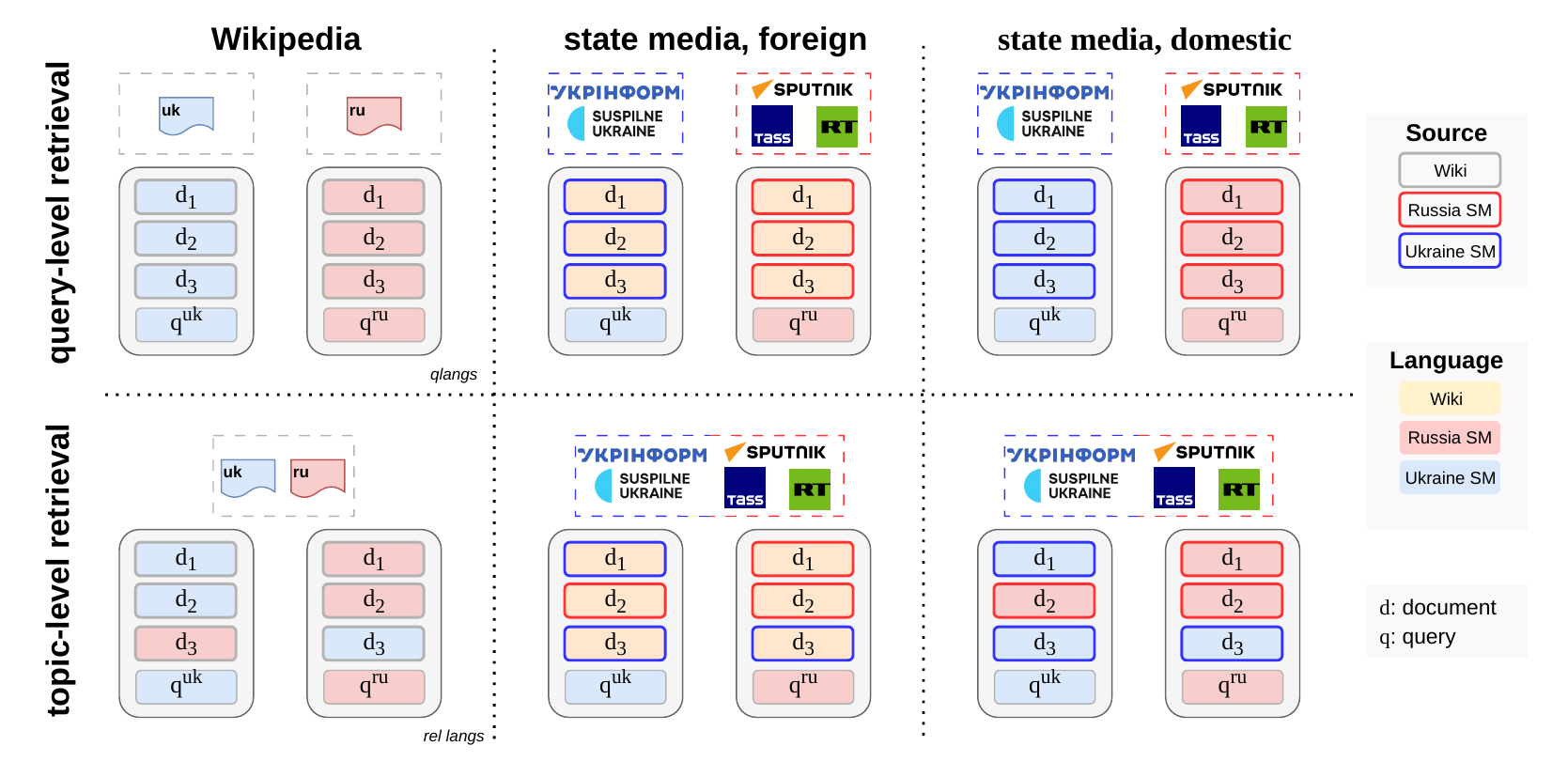}
    \caption[Illustration of the 6 retrieval settings, which vary the sources, and the perspectives.]{Illustration of the 6 retrieval settings studied in \S\ref{sec:rag}, which vary sources (columns) and perspectives (rows). For each box, its fill color indicates the language, while its outline color indicates the source. The Wikipedia settings are single-source, and the state media settings are multi-source.}
    \label{fig:sm_rag}
\end{figure*}

Figure~\ref{fig:sm_rag} illustrates the retrieval settings of Section~\ref{sec:rag}, which vary both sources and perspectives.

\subsection{Concurrence Score Metrics}
The concurrence score (CS) metrics were introduced in~\citet{li2024land}. They measure three aspects of cross-lingual robustness: factuality, consistency, and geopolitical bias. The formulas, which we reproduce from their work, are given in Table~\ref{tab:cs_formulas}.

\label{sec:cs_metrics}
\begin{table*}
    \centering
    \small
    \begin{tabular}{l r@{\;}c@{\;}l l}
\toprule
Full Metric Name & \multicolumn{3}{l}{Formula} & Attribute\\ \midrule
Concurrence Score
  & \csmetric$(c_a, c_b)$
  & $=$
  & $100 \cdot \begin{cases}
        1 & \text{if } c_a = c_b, \\
        0 & \text{otherwise}
     \end{cases}$ & \\
Knowledge-base \csmetric
  & \kbcs$(\mathcal{R}_t)$
  & $=$
  & \csmetric$(c^{KB}, c_\text{en})$  & Factuality\\
Controller \csmetric
  & \concs$(\mathcal{R}_t)$
  & $=$
  & \csmetric$(c^{KB}, c_\text{con})$ & \\
Non-controller \csmetric
  & \noncs$(\mathcal{R}_t)$
  & $=$
  & $\displaystyle\dfrac{1}{|C_{\text{non}}|} \sum_{c\in C_{\text{non}}} \csmetric(c^{KB}, c)$ &\\
Delta \csmetric
  & \deltacs$(\mathcal{R}_t)$
  & $=$
  & $\dfrac{\concs(\mathcal{R}_t) - \noncs(\mathcal{R}_t)}{\noncs(\mathcal{R}_t)}$ & \shortstack{Language Fairness\\ (Geopolitical Bias)} \\
Consistency \csmetric
  & \cstcs$(\mathcal{R}_t)$
  & $=$
  & $\dfrac{1}{n(n-1)} \sum_{i=1}^n \sum_{\substack{j=1, j\ne i}}^n \csmetric(c_i, c_j)$ & Consistency \\
\bottomrule
\end{tabular}

    \caption[Formulas for concurrence score (\csmetric) metrics.]{Formulas for concurrence score (\csmetric) metrics. Each metric is calculated over LLM's multilingual response set for a territory $t$, designated as $\mathcal{R}_t=c_1, ..., c_n$. The response in the controller language is $c_{\text{con}}$, and the set of responses in non-controller languages is $C_{\text{non}}=c_i, ..., c_j$. The gold response is $c^{\text{KB}}$.}
    \label{tab:cs_formulas}
\end{table*}

\section{Additional Results for Case Study}
\label{sec:appendix_pipeline}

\subsection{Additional Tables and Figures}

\begin{figure*}[t]
  \centering
  \begin{minipage}[t]{0.49\textwidth}
    \centering
    \vspace{0pt}
    \includegraphics[width=\linewidth]{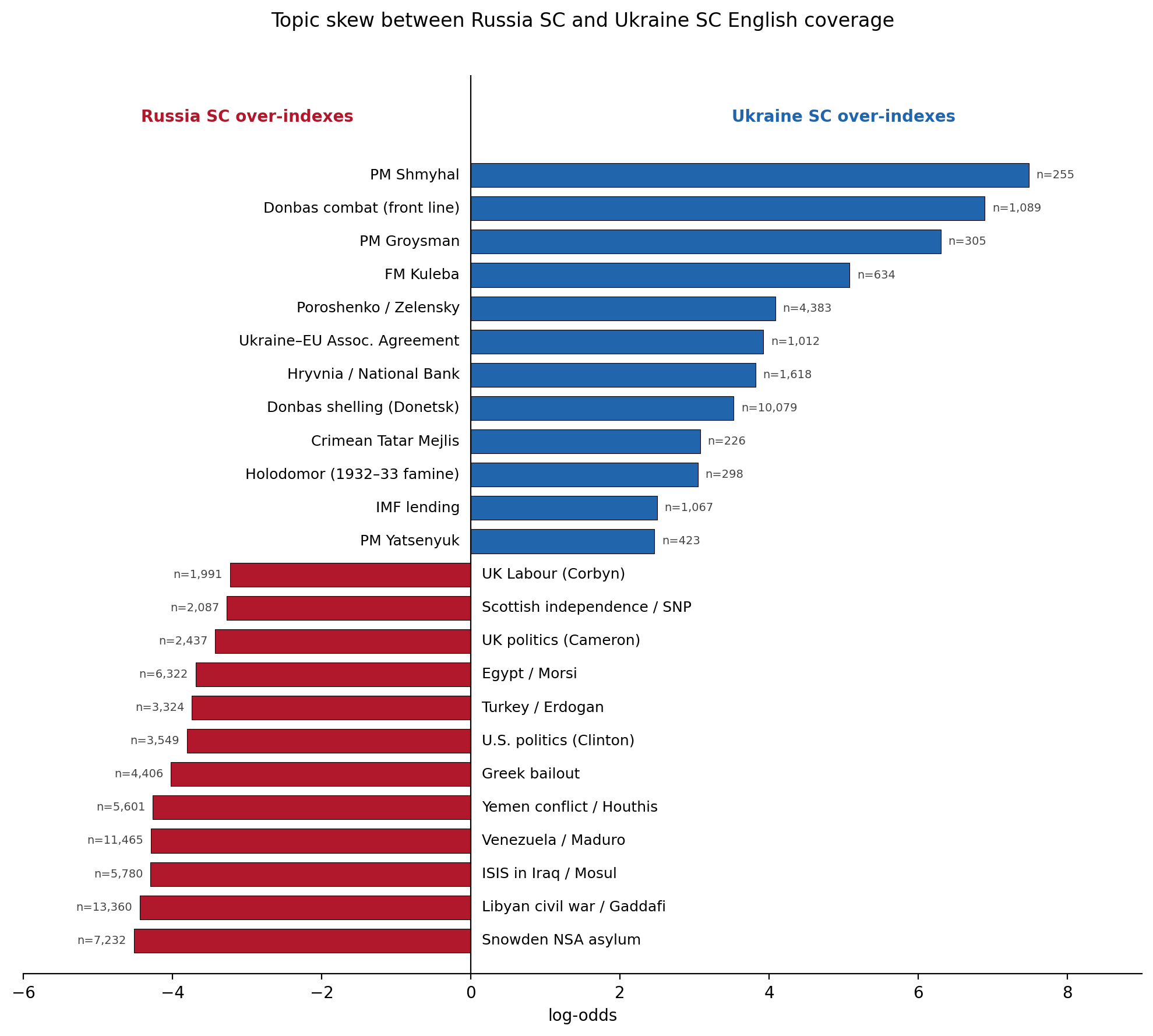}
    \subcaption{English-language media articles}
    \label{fig:topic-skew}
  \end{minipage}
  \hfill
  \begin{minipage}[t]{0.49\textwidth}
    \centering
    \vspace{0pt}
    \includegraphics[width=\linewidth]{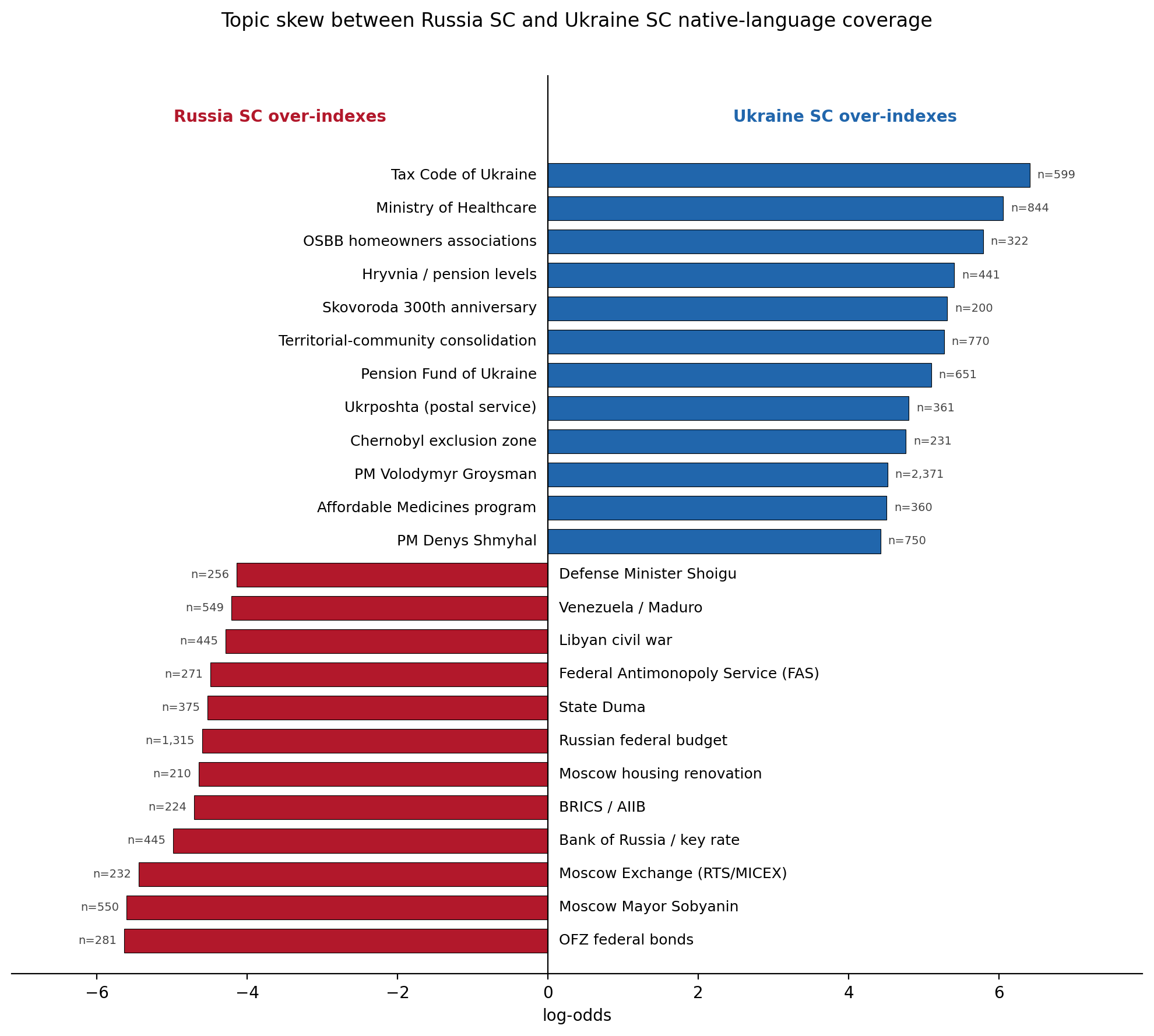}
    \subcaption{Native-language media articles}
    \label{fig:topic-skew-domestic}
  \end{minipage}
  \caption{Topic skew between Russia and Ukraine SC media. Each bar is one fact cluster, positioned by its log-odds of Ukraine-vs.-Russia representation (right/blue over-indexes in Ukraine SC, left/red in Russia SC). $n$ is the number of extracted facts in the cluster (pooled across both sources), a measure of topic volume that is independent of the bar length.}
  \label{fig:topic-skew-combined}
\end{figure*}

To complement the case study in \S\ref{sec:case-study}, Figures~\ref{fig:topic-skew} and~\ref{fig:topic-skew-domestic} present the top Russia-vs.-Ukraine topic-skew distributions for the English-language and native-language corpora.



We include the statistics of the Russian and Ukrainian SC corpora by outlet, broken down into English and native-language splits in Table~\ref{tab:corpus-stats}.

\subsection{LLM-based Analysis of State Media Topics on Crimea}
\label{sec:crimea_llm_analysis}
For additional large-scale and multilingual analysis, we apply a blind LLM-agent comparison to the Crimea-related topics in both the English-language and native-language corpora. Claude Sonnet 4.6 analyzes each topic with country labels omitted.

\paragraph{English-language extracted topics}
On the ten English-language Crimea topics, the blind LLM analysis reports two incompatible narratives on the same events. On the referendum topic, Russia SC presents 96--97\% support for ``reunification,'' emphasizing voter turnout and procedural detail. Ukraine SC frames the same event as an ``illegitimate annexation'' conducted ``under pressure from Russian military.'' The Kerch Strait Bridge is described in Russia SC via technical specifications and construction milestones, and in Ukraine SC as connecting to ``illegally-annexed'' territory. Aksyonov appears in Russia SC under his official title and reported policy initiatives, and in Ukraine SC as head of a ``puppet government'' in ``Russian-controlled'' Crimea. The Crimean Tatar Mejlis appears in Russia SC as the subject of extremism allegations and a legal ban, and in Ukraine SC as legitimate representatives under occupation.

\paragraph{Native-language extracted topics}
On the six native-language Crimea topics, the LLM analysis reports consistent framing differences. On the 2014 referendum, Russia SC reports 96.77\% support and historical Russian ties, while Ukraine SC frames the event as a violation of international law. On the Kerch Strait Bridge, Russia SC reports construction milestones and economic benefits, while Ukraine SC reports archaeological disruption and economic damage to shipping. On the Crimean Tatar Mejlis, Russia SC questions its legitimacy and links its leaders to extremism, while Ukraine SC emphasizes Russian persecution and cites UN and ICJ rulings. On the water-supply dispute, Russia SC documents Crimea's water shortage after the North Crimean Canal closure and Russian desalination plans, while Ukraine SC focuses on mainland infrastructure.
Tables~\ref{tab:example-facts} and~\ref{tab:example-facts-domestic} show representative extracted facts from Crimea topics that complement the analysis.

\begin{table*}[t]
\centering
\small
\renewcommand{\arraystretch}{1.25}
\begin{tabularx}{\textwidth}{@{}p{0.47\textwidth} p{0.47\textwidth}@{}}
\toprule
\textbf{Russia SC} & \textbf{Ukraine SC} \\
\midrule
\multicolumn{2}{@{}l}{\textit{2014 Crimean referendum}} \\
\addlinespace[2pt]
[claim] The Crimea peninsula rejoined Russia in March. &
[claim] Russia annexed Ukraine's Crimea peninsula in early 2014. \\
\addlinespace[3pt]
[claim] The Crimean parliament voted overwhelmingly to hold a referendum. &
[claim] The March 16, 2014 referendum in Crimea and Sevastopol was conducted under pressure from the Russian military and with the support of Russian propagandist media. \\
\midrule
\multicolumn{2}{@{}l}{\textit{Crimean Tatar Mejlis}} \\
\addlinespace[2pt]
[attribution] Crimea's Prosecutor Natalia Poklonskaya warned Refat Chubarov early in July that the Crimean authorities might recognize the Crimean Parliament (Mejlis) to be an extremist organization. &
[attribution] Refat Chubarov is the Head of the Mejlis of the Crimean Tatar people and a Member of Parliament. \\
\addlinespace[3pt]
[claim] Refat Chubarov, the chief of the Crimean Parliament (Mejlis), has been barred from entering Crimea for five years. &
[attribution] Deputy Chairman of the Mejlis of the Crimean Tatar people Akhtem Chiygoz stated that the Russian Federation is conducting state terrorism in Crimea against the indigenous population. \\
\midrule
\multicolumn{2}{@{}l}{\textit{November 2018 Kerch Strait incident}} \\
\addlinespace[2pt]
[claim] Ukraine's Berdyansk and Nikopol gunboats and the Yany Kapu tugboat illegally traversed the Russian maritime border. &
[claim] Russian border ships fired on and seized three Ukrainian vessels on November 25, 2018, in the Kerch Strait while the vessels were traveling from the port of Odesa to the port of Mariupol. \\
\addlinespace[3pt]
[statistic] Over 20 Ukrainian sailors were restrained when the vessels were arrested by Russia. &
[statistic] Twenty-four Ukrainian sailors were captured when Russian border ships seized three Ukrainian vessels on November 25, 2018, in the Kerch Strait. \\
\bottomrule
\end{tabularx}
\caption{Representative decontextualized facts extracted from Russia and Ukraine SC English corpus on three Crimea related topics. Bracketed tags represent the automatically assigned fact categories from our pipeline. The two sources often cover the same events with divergent terminology and framing.}
\label{tab:example-facts}
\end{table*}

\begin{table*}[t]
\centering
\small
\renewcommand{\arraystretch}{1.25}
\begin{tabularx}{\textwidth}{@{}p{0.47\textwidth} p{0.47\textwidth}@{}}
\toprule
\textbf{Russia SC} & \textbf{Ukraine SC} \\
\midrule
\multicolumn{2}{@{}l}{\textit{2014 Crimean referendum}} \\
\addlinespace[2pt]
[claim] Crimea became a Russian region after a referendum held in March 2014. &
[claim] Crimea, along with the ``LNR'' and ``DNR,'' are considered temporarily occupied territories of Ukraine. \\
\addlinespace[3pt]
[statistic] In the March 2014 referendum, 96.77\% of voters in the Republic of Crimea and 95.6\% of residents in Sevastopol voted to join Russia. &
[claim] The annexation of Crimea and the subsequent attempt to create ``Novorossiya'' are described as brazen actions by Russia. \\
\midrule
\multicolumn{2}{@{}l}{\textit{Crimean Tatar Mejlis}} \\
\addlinespace[2pt]
[attribution] Poklonskaya stated that the Mejlis exists to incite the population, commit acts of sabotage, promote war, incite hatred, and disgrace the Crimean Tatar people and their symbols. &
[attribution] Refat Chubarov, the head of the Mejlis of the Crimean Tatar people, believes the UN resolution's importance lies in enabling Ukraine to demand that specialized structures and international organizations send monitoring missions to the peninsula. \\
\addlinespace[3pt]
[claim] Refat Chubarov and Mustafa Dzhemilev were banned from entering Russia for five years starting in 2014 due to extremist statements that incited national discord. &
[claim] Russia must immediately revoke the decision to declare the Mejlis of the Crimean Tatar People an extremist organization and ban its activities, and also annul the decision to prohibit Mejlis leaders from entering Crimea. \\
\midrule
\multicolumn{2}{@{}l}{\textit{November 2018 Kerch Strait incident}} \\
\addlinespace[2pt]
[claim] On November 25, 2018, Ukrainian Navy ships ``Berdyansk,'' ``Nikopol,'' and ``Yany Kapu'' crossed the state border of Russia in violation of Articles 19 and 21 of the United Nations Convention on the Law of the Sea. &
[claim] Russian military forces fired upon and seized the vessels ``Berdyansk'' and ``Nikopol,'' along with the tugboat ``Yany Kapu,'' and their crews in the Kerch Strait. \\
\addlinespace[3pt]
[attribution] The Federal Security Service (FSB) of Russia stated that the Ukrainian Navy ships ``Berdyansk,'' ``Nikopol,'' and ``Yany Kapu'' maneuvered dangerously before being detained. &
[statistic] Twenty-four Ukrainians were accused of illegally crossing the Russian border (Part 3 of Article 322 of the Criminal Code of Russia), facing a sentence of up to six years in prison. \\
\bottomrule
\end{tabularx}
\caption{Representative decontextualized facts extracted from Russia and Ukraine native-language \smsc\ media on three Crimea topics. Bracketed tags represent the auto-assigned fact categories from our pipeline. The two sources often cover the same events with divergent terminology and framing.}
\label{tab:example-facts-domestic}
\end{table*}

\begin{table}[t]
  \centering
  \small
  \begin{tabular}{l r r r}
    \toprule
    Outlet & Foreign & Domestic & Total \\
           & (EN)    & (native) &       \\
    \midrule
    \multicolumn{4}{l}{\textit{Russia SC}} \\
    rt.com            & 105{,}579 &      -- & 105{,}579 \\
    tass.com          &  50{,}191 &       1 &  50{,}192 \\
    sputniknews.com   &  43{,}491 &       1 &  43{,}492 \\
    rbth.com          &  15{,}897 &       1 &  15{,}898 \\
    ria.ru            &        -- &  8{,}473 &   8{,}473 \\
    rg.ru             &        -- &  6{,}253 &   6{,}253 \\
    m24.ru            &        -- &  2{,}421 &   2{,}421 \\
    mir24.tv          &        -- &  1{,}989 &   1{,}989 \\
    inosmi.ru         &        -- &  1{,}802 &   1{,}802 \\
    rsport.ria.ru     &        -- &  1{,}762 &   1{,}762 \\
    vesti.ru          &       22 &  1{,}692 &   1{,}714 \\
    otr-online.ru     &        -- &  1{,}567 &   1{,}567 \\
    1prime.ru         &        -- &  1{,}263 &   1{,}263 \\
    crimea.ria.ru     &        -- &  1{,}034 &   1{,}034 \\
    tvzvezda.ru       &        1 &     972 &     973 \\
    realty.ria.ru     &        -- &     922 &     922 \\
    rtd.rt.com        &      790 &      -- &     790 \\
    smotrim.ru        &        9 &     723 &     732 \\
    mirtv.ru          &        -- &     103 &     103 \\
    Other (4 outlets) &        4 &      21 &      25 \\
    \addlinespace
    \textbf{Russia subtotal} & \textbf{215{,}984} & \textbf{31{,}000} & \textbf{246{,}984} \\
    \midrule
    \multicolumn{4}{l}{\textit{Ukraine SC}} \\
    ukurier.gov.ua    &        -- & 13{,}093 &  13{,}093 \\
    golos.com.ua      &       17 &  4{,}858 &   4{,}875 \\
    ukrinform.net     &   4{,}049 &      -- &   4{,}049 \\
    \addlinespace
    \textbf{Ukraine subtotal} & \textbf{4{,}066} & \textbf{17{,}951} & \textbf{22{,}017} \\
    \midrule
    \textbf{Total} & \textbf{220{,}050} & \textbf{48{,}951} & \textbf{269{,}001} \\
    \bottomrule
  \end{tabular}
  \caption{Corpus composition by outlet and split. Counts are URL-deduplicated articles. Foreign/domestic split is based on the article's publication language (English $\rightarrow$ foreign-targeted, Russian or Ukrainian $\rightarrow$ domestic). Outlets with fewer than 20 articles are grouped into ``Other.''}
  \label{tab:corpus-stats}
\end{table}

\section{Details on Fact Extraction}
\label{sec:fact_extraction}
The decontextualized fact extraction pipeline is inspired by~\citet{choi-etal-2021-decontextualization}, but implemented with modern LLMs. We detail the extraction, clustering, and analysis steps below; all steps are run on a single NVIDIA B200 GPU.

\subsection{Implementation Details}
\label{sec:fact_extraction_details}
Facts in the English split are extracted with a Qwen3-4B model distilled from Claude Sonnet~4.6 and served with vLLM, using greedy decoding with a $16{,}384$-token context and up to $8{,}192$ generated tokens per article. Facts in the Russian and Ukrainian (native-language) split are extracted with Gemini 2.5 Flash-Lite using the same prompt with an instruction to generate English-translated facts. Given each news article, the model generated a decontextualized version of the facts so that they can be understood without the source article. Figure~\ref{fig:fact_prompt} shows the prompt used for fact extraction.

\paragraph{Embedding}
Each fact's decontextualized text is embedded with BGE-M3 embedding \citep{chen-etal-2024-m3} using the sentence-transformers library \citep{reimers-2019-sentence-bert}, producing $1{,}024$-dimensional $L_2$-normalized vectors. We embed the Russia and Ukraine fact sets jointly, retaining a country label per fact.

\paragraph{Clustering}
We reduce the $1{,}024$-dimensional embeddings to 5 dimensions with UMAP (\texttt{n\_neighbors}=15, \texttt{min\_dist}=0) and cluster the reduced space with HDBSCAN (\texttt{min\_cluster\_size}=200, \texttt{min\_samples}=20). We label each cluster with its most characteristic terms using class-based TF-IDF , where per-cluster term frequencies are normalized by cluster size and weighted by inverse cluster frequency. 

\subsection{Topic Framing Analysis}
\label{sec:topic_framing_analysis}
For each cluster we compute a Laplace-smoothed log-odds ratio of Ukraine versus Russia representation to rank topics by asymmetric coverage. For topics covered by both corpora, we additionally run a blind LLM-agent comparison using Claude Sonnet~4.6. For each topic, the facts from Russia SC and Ukraine SC are presented to the model with the two sources anonymized (relabeled as ``Source A'' and ``Source B'') and randomly ordered, and the model produces a per-topic analysis of framing, sourcing, and terminology differences.

\begin{figure*}[ht]
    \centering
    \begin{tcolorbox}[enhanced, width=\linewidth, boxrule=0.8mm]
        \footnotesize
        \begin{verbatim}
You are a fact extraction system. Given a news article, extract factual content, separating it
from expressive/creative/writerly content.

YOUR #1 PRIORITY is thorough decontextualization. Every fact you output must be rewritten so
a reader with ZERO context can understand it: resolve ALL pronouns and definite references
to the specific named entity, expand ALL abbreviations, add geographic/institutional context
that is stated in the article, and reproduce all numbers exactly. Do NOT infer or add
information beyond what the article states (e.g. do not add a year that is not present).

EXTRACT: findings, data points and statistics, named entities, dates, methodology,
causal claims, attributions, and relationships (who funds / owns / manages / tests what).
DISCARD (into a "discarded" list, with a reason): opinions, speculation, hyperbole, predictions,
rhetorical questions, narrative transitions, and editorial commentary.

[... Detailed decontextualization rules with worked BAD/GOOD examples omitted ...]

CATEGORIZE each fact as one of:
finding | methodology | background | attribution | statistic | claim
TAG each fact's temporal scope: timeless | time-bound | event-specific
(with a temporal_anchor date/period for time-bound and event-specific facts).
TAG each fact's evidence_type: quoted (named person) | study | editorial | established.
Set a confidence score in [0, 1].

Return ONLY a valid JSON object:
{
  "key_claims": ["A decontextualized paragraph-level claim summarizing a main point."],
  "facts": [
    {"text": "The decontextualized factual statement.",
     "original_sentence": "The original sentence from the article.",
     "category": "finding", "temporal": "time-bound", "temporal_anchor": "January 2025",
     "evidence_type": "study", "confidence": 0.95, "supports_claims": [0]}
  ],
  "discarded": [{"text": "The discarded text.", "reason": "opinion"}],
  "publication_date": "YYYY-MM-DD only if explicitly stated, otherwise empty string"
}
\end{verbatim}
    \end{tcolorbox}
    \caption{Prompt for decontextualized fact extraction.}
    \label{fig:fact_prompt}
\end{figure*}

\section{Implementation Details for Information Pursuit}
\label{sec:appe_information_pursuit}

Table~\ref{tab:ip_query_bank} lists the full set of binary yes/no queries used in the Information Pursuit analysis of Section~\ref{sec:ip_method}. Each query targets a specific narrative framing dimension and is answered by the LLM answerer for each input article. We use \texttt{GPT-5-mini} as the generator of the query bank and query answerer. We use \texttt{Llama-3.1-8B-Instruct} as the posterior predictor, with $M=16$ samples per iteration of query selection and $T=10$  iterations per document.

\paragraph{Query selection}
IP classifies a document by sequentially picking concept queries $q$, obtaining the grounded answers $a$, and accumulating the history $h_t = \{(q_1, a_1), \ldots, (q_{t-1}, a_{t-1})\}$ as evidence for a posterior predictor. At iteration $t$ with accumulated history $h_t$, for each candidate query $q$ we sample $M$ calibration examples. For the $m$-th calibration example we simulate $h_t^{(m)} = h_t \cup \{(q, a_q^{(m)})\}$ and obtain the posterior $p(y \mid h_t^{(m)})$ from the LLM predictor. The expected posterior entropy for $q$ is
\begin{equation*}
\scriptstyle
    \hat{H}(q) = \frac{1}{M}\sum_{m=1}^{M}\!\left(-\sum_y p\!\left(y \mid h_t^{(m)}\right)\log p\!\left(y \mid h_t^{(m)}\right)\right),
    \label{eq:ip_entropy}
\end{equation*}
and the next most informative query is selected to minimize the posterior entropy
\begin{equation*}
    q_{t+1} = \arg\min_{q \in \mathcal{Q}_t} \hat{H}(q),
    \label{eq:ip_selection}
\end{equation*}
which approximates the query selection process that maximizes the mutual information.

\begin{figure}[t]
  \centering
  \includegraphics[width=\linewidth]{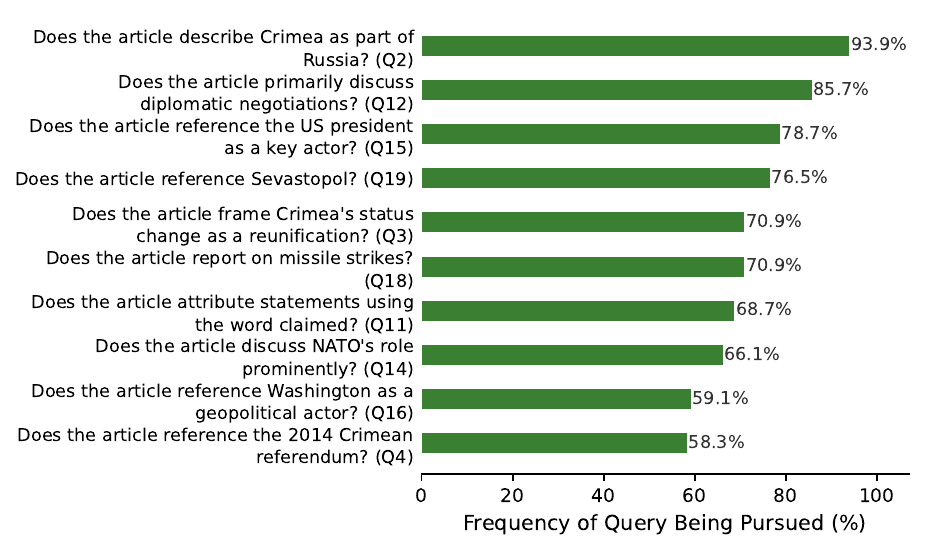}
  \caption{For articles on topic Crimea, frequency of each query being pursued by the information pursuit system (query descriptions are in Table~\ref{tab:ip_query_bank}).}
  \label{fig:ip_queries}
\end{figure}

Figure~\ref{fig:ip_queries} shows how often each query is selected across the Crimea articles.

\begin{table}[t!]
    \centering
    \small
    \begin{tabularx}{\linewidth}{ccX}
        \toprule
        Category & ID & Query \\
        \midrule
        \multirow{8}{*}{\rotatebox[origin=c]{90}{\begin{tabular}{@{}c@{}}Territorial Status \\ Framing\end{tabular}}}
            & Q1  & Does the article describe Crimea as occupied territory? \\
            & Q2  & Does the article describe Crimea as part of Russia? \\
            & Q3  & Does the article frame Crimea's status change as a reunification? \\
            & Q4  & Does the article reference the 2014 Crimean referendum? \\
        \midrule
        \multirow{6}{*}{\rotatebox[origin=c]{90}{\begin{tabular}{@{}c@{}}Agency \\ Attribution\end{tabular}}}
            & Q5  & Does the article frame Russia as the aggressor in the conflict? \\
            & Q6  & Does the article report Ukrainian military successes against Russian targets? \\
            & Q7  & Does the article report Ukrainian strikes on targets in Crimea? \\
        \midrule
        \multirow{4}{*}{\rotatebox[origin=c]{90}{\begin{tabular}{@{}c@{}}Source \\ Authority\end{tabular}}}
            & Q8  & Does the article quote Russian government officials as primary sources? \\
            & Q9  & Does the article reference Moscow as a decision-making authority? \\
        \midrule
        \multirow{8}{*}{\rotatebox[origin=c]{90}{\begin{tabular}{@{}c@{}}Rhetorical \\ Style\end{tabular}}}
            & Q10 & Does the article use hedging language such as reportedly or allegedly? \\
            & Q11 & Does the article attribute statements using the word claimed? \\
            & Q12 & Does the article primarily discuss diplomatic negotiations? \\
            & Q13 & Does the article emphasize historical context preceding the conflict? \\
        \midrule
        \multirow{6}{*}{\rotatebox[origin=c]{90}{\begin{tabular}{@{}c@{}}External Actor \\ Focus\end{tabular}}}
            & Q14 & Does the article discuss NATO's role prominently? \\
            & Q15 & Does the article reference the US president as a key actor? \\
            & Q16 & Does the article reference Washington as a geopolitical actor? \\
        \midrule
        \multirow{5}{*}{\rotatebox[origin=c]{90}{\begin{tabular}{@{}c@{}}Geographic \\ Focus\end{tabular}}}
            & Q17 & Does the article discuss events in the Donbas region? \\
            & Q18 & Does the article report on missile strikes? \\
            & Q19 & Does the article reference Sevastopol? \\
        \bottomrule
    \end{tabularx}
    \caption{Full query bank used in the Information Pursuit algorithm of Crimea-related articles. Each row corresponds to one binary yes/no question posed to the LLM answerer. The Category column groups queries by the narrative framing dimension they target.}
    \label{tab:ip_query_bank}
\end{table}

\end{document}